\documentclass[sigconf,dvipsnames]{acmart}
\usepackage[noadjust]{cite}
\usepackage[font=footnotesize]{subcaption}
\usepackage[font=footnotesize]{caption}
\usepackage{caption}
\usepackage[utf8]{inputenc}
\usepackage{graphicx,url}

\usepackage{booktabs}
\usepackage{diagbox}
\graphicspath{ {./images/} }
\usepackage{multirow}
\usepackage{hhline}
\usepackage{xspace}
\usepackage{graphicx}
\usepackage{amsmath,mathtools}
\usepackage{amsthm}
\newtheorem{theorem}{Theorem}   %
\newtheorem{proposition}{Proposition} %
\newtheorem{remark}{Remark}
\newtheorem{lemma}{Lemma}

\usepackage{bm}                 %
\usepackage{pifont}

\newtheorem{assumption}{Assumption}
\theoremstyle{plain}

\usepackage{color,soul}
\usepackage{xcolor}
\usepackage{bbm}
\usepackage{stmaryrd}
\usepackage{textcomp}
\usepackage{lipsum,lineno}
\usepackage{float}
\makeatletter
\newif\if@restonecol
\makeatother
\usepackage{amsmath}
\usepackage{mathrsfs}
\usepackage[shortlabels]{enumitem}
\newcommand\numeq[1]%
{\stackrel{\scriptscriptstyle(\mkern-1.5mu#1\mkern-1.5mu)}{=}}
\newcommand\numeqq[1]%
{\stackrel{\scriptscriptstyle(\mkern-1.5mu#1\mkern-1.5mu)}{\triangleq}}
\newcommand\numleq[1]%
{\stackrel{\scriptscriptstyle(\mkern-1.5mu#1\mkern-1.5mu)}{\leq}}
\newcommand\numgeq[1]%
{\stackrel{\scriptscriptstyle(\mkern-1.5mu#1\mkern-1.5mu)}{\geq}}
\newcommand\numimp[1]%
{\stackrel{\scriptscriptstyle(\mkern-1.5mu#1\mkern-1.5mu)}{\implies}}
\newcommand\norm[1]{\lVert#1\rVert}

\usepackage[titlenumbered,ruled,linesnumbered]{algorithm2e}
\SetAlCapNameFnt{\footnotesize}
\SetAlCapFnt{\footnotesize}

\let\oldnl\nl% Store \nl in \oldnl
\newcommand{\nonl}{\renewcommand{\nl}{\let\nl\oldnl}}
\usepackage{algpseudocode}

\usepackage{fancyhdr}
\usepackage{tabularx}
\usepackage{dsfont}
\usepackage{listings}
\usepackage[most]{tcolorbox}
\lstdefinestyle{mystyle}{
    backgroundcolor=\color{backcolour},
    commentstyle=\color{codegreen},
    keywordstyle=\color{magenta},
    numberstyle=\tiny\color{codegray},
    stringstyle=\color{codepurple},
    basicstyle=\ttfamily\footnotesize,
    breakatwhitespace=false,
    breaklines=true,
    captionpos=b,
    keepspaces=true,
    showspaces=false,
    showstringspaces=false,
    showtabs=false,
    tabsize=2,
    xleftmargin=50pt,
    xrightmargin=50pt
  }
\usepackage{tikz,pgfplots}
\usepackage{pgfplots}
\usepackage{pgfplotstable}
\definecolor{gray2}{HTML}{ededed}
\definecolor{gray3}{HTML}{F5F5F5}
\definecolor{RoyalAzure}{rgb}{0.0, 0.22, 0.66}
\definecolor{lightgray}{gray}{0.9}
\definecolor{lightgray}{gray}{0.9}
\definecolor{lightgreen}{rgb}{0.88, 1, 0.88}
\definecolor{lightred}{rgb}{1, 0.88, 0.88}
\definecolor{lightblue}{rgb}{0.88, 0.94, 1}
\usepackage{colortbl}
\usetikzlibrary{pgfplots.groupplots}
\usepgfplotslibrary{colorbrewer}
\pgfplotsset{compat = 1.15, cycle list/Set1-8}
\usetikzlibrary{pgfplots.statistics, pgfplots.colorbrewer}
\usetikzlibrary{pgfplots.groupplots}
\usetikzlibrary{shapes.geometric,backgrounds,patterns, trees}
\usetikzlibrary{3d,decorations.text,shapes.arrows,positioning,fit,backgrounds}
\usetikzlibrary{positioning, decorations.pathmorphing, shapes}
\usetikzlibrary{decorations.pathreplacing}
\usetikzlibrary{shapes.geometric,backgrounds,patterns, trees}
\usetikzlibrary{spy}
\usetikzlibrary{arrows.meta,
                bending,
                intersections,
                quotes,
                shapes.geometric}
              \usetikzlibrary{automata, positioning}
              \usepgfplotslibrary{fillbetween}
\usetikzlibrary{shapes,arrows}
\usetikzlibrary{arrows.meta}
\usetikzlibrary{positioning}
\tikzset{set/.style={draw,circle,inner sep=0pt,align=center}}
\usetikzlibrary{automata, positioning}
  \usetikzlibrary{shapes,shadows}
  \tikzstyle{abstractbox} = [draw=black, fill=white, rectangle,
  inner sep=10pt, style=rounded corners, drop shadow={fill=black,
  opacity=1}]
\tikzstyle{abstracttitle} =[fill=white]
\usetikzlibrary{calc,positioning,shapes.geometric}
\usetikzlibrary{arrows.meta,arrows}

\DeclareMathOperator*{\argmin}{arg\,min}

\DeclareMathOperator*{\minimize}{minimize}

\usetikzlibrary{matrix}
\tikzstyle{cblue}=[circle, draw, thin,fill=cyan!20, scale=0.8]
\tikzstyle{qgre}=[rectangle, draw, thin,fill=green!20, scale=0.8]
\tikzstyle{rpath}=[ultra thick, red, opacity=0.4]
\tikzstyle{legend_isps}=[rectangle, rounded corners, thin,
                       fill=gray!20, text=blue, draw]

\tikzstyle{legend_overlay}=[rectangle, rounded corners, thin,
                           top color= white,bottom color=green!25,
                           minimum width=2.5cm, minimum height=0.8cm,
                           pinegreen]
\tikzstyle{legend_phytop}=[rectangle, rounded corners, thin,
                          top color= white,bottom color=cyan!25,
                          minimum width=2.5cm, minimum height=0.8cm,
                          royalblue]
\tikzstyle{legend_general}=[rectangle, rounded corners, thin,
                          top color= white,bottom color=lavander!25,
                          minimum width=2.5cm, minimum height=0.8cm,
                          violet]
                          \usetikzlibrary{matrix}

\colorlet{myRed}{red!20}
\tikzset{
  rows/.style 2 args={/utils/temp/.style={row ##1/.append style={nodes={#2}}},
    /utils/temp/.list={#1}},
  columns/.style 2 args={/utils/temp/.style={column ##1/.append style={nodes={#2}}},
    /utils/temp/.list={#1}}}
\usetikzlibrary{backgrounds,calc,shadings,shapes.arrows,shapes.symbols,shadows}
\definecolor{switch}{HTML}{006996}

\makeatletter
\pgfkeys{/pgf/.cd,
  parallelepiped offset x/.initial=2mm,
  parallelepiped offset y/.initial=2mm
}
\pgfdeclareshape{parallelepiped}
{
  \inheritsavedanchors[from=rectangle] % this is nearly a rectangle
  \inheritanchorborder[from=rectangle]
  \inheritanchor[from=rectangle]{north}
  \inheritanchor[from=rectangle]{north west}
  \inheritanchor[from=rectangle]{north east}
  \inheritanchor[from=rectangle]{center}
  \inheritanchor[from=rectangle]{west}
  \inheritanchor[from=rectangle]{east}
  \inheritanchor[from=rectangle]{mid}
  \inheritanchor[from=rectangle]{mid west}
  \inheritanchor[from=rectangle]{mid east}
  \inheritanchor[from=rectangle]{base}
  \inheritanchor[from=rectangle]{base west}
  \inheritanchor[from=rectangle]{base east}
  \inheritanchor[from=rectangle]{south}
  \inheritanchor[from=rectangle]{south west}
  \inheritanchor[from=rectangle]{south east}
  \backgroundpath{
    \southwest \pgf@xa=\pgf@x \pgf@ya=\pgf@y
    \northeast \pgf@xb=\pgf@x \pgf@yb=\pgf@y
    \pgfmathsetlength\pgfutil@tempdima{\pgfkeysvalueof{/pgf/parallelepiped
      offset x}}
    \pgfmathsetlength\pgfutil@tempdimb{\pgfkeysvalueof{/pgf/parallelepiped
      offset y}}
    \def\ppd@offset{\pgfpoint{\pgfutil@tempdima}{\pgfutil@tempdimb}}
    \pgfpathmoveto{\pgfqpoint{\pgf@xa}{\pgf@ya}}
    \pgfpathlineto{\pgfqpoint{\pgf@xb}{\pgf@ya}}
    \pgfpathlineto{\pgfqpoint{\pgf@xb}{\pgf@yb}}
    \pgfpathlineto{\pgfqpoint{\pgf@xa}{\pgf@yb}}
    \pgfpathclose
    \pgfpathmoveto{\pgfqpoint{\pgf@xb}{\pgf@ya}}
    \pgfpathlineto{\pgfpointadd{\pgfpoint{\pgf@xb}{\pgf@ya}}{\ppd@offset}}
    \pgfpathlineto{\pgfpointadd{\pgfpoint{\pgf@xb}{\pgf@yb}}{\ppd@offset}}
    \pgfpathlineto{\pgfpointadd{\pgfpoint{\pgf@xa}{\pgf@yb}}{\ppd@offset}}
    \pgfpathlineto{\pgfqpoint{\pgf@xa}{\pgf@yb}}
    \pgfpathmoveto{\pgfqpoint{\pgf@xb}{\pgf@yb}}
    \pgfpathlineto{\pgfpointadd{\pgfpoint{\pgf@xb}{\pgf@yb}}{\ppd@offset}}
  }
}

\makeatletter
\tikzset{anchor/.append code=\let\tikz@auto@anchor\relax,
  add font/.code=%
    \expandafter\def\expandafter\tikz@textfont\expandafter{\tikz@textfont#1},
  left delimiter/.style 2 args={append after command={\tikz@delimiter{south east}
    {south west}{every delimiter,every left delimiter,#2}{south}{north}{#1}{.}{\pgf@y}}}}
\tikzstyle{sms} = [rectangle callout, draw,very thick, rounded corners, minimum height=20pt]
\makeatletter
\tikzset{anchor/.append code=\let\tikz@auto@anchor\relax,
  add font/.code=%
    \expandafter\def\expandafter\tikz@textfont\expandafter{\tikz@textfont#1},
  left delimiter/.style 2 args={append after command={\tikz@delimiter{south east}
    {south west}{every delimiter,every left delimiter,#2}{south}{north}{#1}{.}{\pgf@y}}}}
\tikzstyle{sms} = [rectangle callout, draw,very thick, rounded corners, minimum height=20pt]
\usetikzlibrary{positioning,calc}
\tikzstyle{block} = [rectangle, draw,
text width=10.5em, text centered, rounded corners, minimum height=4em]
\tikzstyle{line} = [draw, -latex]
\tikzset{
  mybackground51/.style={execute at end picture={
      \begin{scope}[on background layer]
        \draw[black, rounded corners=2ex, fill=gray2] (current bounding box.south west)
        rectangle (current bounding box.north east);
        \node[draw,fill=white,ellipse,anchor=west,inner sep=1pt,minimum width=1ex] at (current bounding box.north
        west){#1};
      \end{scope}
    }},
}
\tikzset{
  mybackground9/.style={execute at end picture={
        \begin{scope}[on background layer]
          \draw[black,fill=black!5,rounded corners=6ex] (current bounding box.south west)
                    rectangle (current bounding box.north east);
          \node[draw,fill=white,ellipse,anchor=west,inner sep=1pt,minimum width=4ex] at (current bounding box.north
                   west){#1};
        \end{scope}
    }},
}

\tikzset{
  mybackground13/.style={execute at end picture={
        \begin{scope}[on background layer]
          \draw[black, fill=gray2, rounded corners=4ex] (current bounding box.south west)
                    rectangle (current bounding box.north east);
          \node[draw,fill=white,ellipse,anchor=west,inner sep=1pt,minimum width=4ex] at (current bounding box.north
                   west){#1};
        \end{scope}
    }},
}
\tikzset{
  mybackground14/.style={execute at end picture={
        \begin{scope}[on background layer]
          \draw[black, rounded corners=2ex] (current bounding box.south west)
                    rectangle (current bounding box.north east);
          \node[draw,fill=white,ellipse,anchor=west,inner sep=1pt,minimum width=4ex] at (current bounding box.north
                   west){#1};
        \end{scope}
    }},
}

\tikzset{
  mybackground6/.style={execute at end picture={
        \begin{scope}[on background layer]
          \draw[black,rounded corners=1ex, line width=0.15mm] (current bounding box.south west)
                    rectangle (current bounding box.north east);
          \node[draw,fill=white,ellipse,anchor=west,inner sep=1pt,minimum width=4ex] at (current bounding box.north
                   west){#1};
        \end{scope}
c    }},
}

\tikzset{
  mybackground11/.style={execute at end picture={
        \begin{scope}[on background layer]
          \draw[black, fill=Black!80!Sepia!9, rounded corners=6ex] (current bounding box.south west)
                    rectangle (current bounding box.north east);
          \node[draw,fill=white,ellipse,anchor=west,inner sep=1pt,minimum width=4ex] at (current bounding box.north
                   west){#1};
        \end{scope}
    }},
}

\tikzset{
  mybackground15/.style={execute at end picture={
        \begin{scope}[on background layer]
          \draw[black, fill=Black!80!Sepia!9, rounded corners=3ex] (current bounding box.south west)
                    rectangle (current bounding box.north east);
          \node[draw,fill=white,ellipse,anchor=west,inner sep=1pt,minimum width=4ex] at (current bounding box.north
                   west){#1};
        \end{scope}
    }},
}

\tikzset{
  mybackground12/.style={execute at end picture={
        \begin{scope}[on background layer]
          \draw[black, fill=Black!40!Emerald!30, rounded corners=3ex, line width=0.3mm] (current bounding box.south west)
                    rectangle (current bounding box.north east);
        \end{scope}
    }},
}
\tikzset{
  mybackground18/.style={execute at end picture={
      \begin{scope}[on background layer]
        \draw[black, fill=gray3, rounded corners=3.5ex] (current bounding box.south west)
        rectangle (current bounding box.north east);
        \node[draw,fill=white,ellipse,anchor=west,inner sep=1pt,minimum width=4ex] at (current bounding box.north
        west){#1};
      \end{scope}
    }}
}
\tikzset{
  mybackground58/.style={execute at end picture={
        \begin{scope}[on background layer]
          \draw[black, fill=blue!40!black!5, rounded corners=1ex] (current bounding box.south west)
                    rectangle (current bounding box.north east);
          \node[draw,fill=white,ellipse,anchor=west,inner sep=1pt,minimum width=4ex, rounded corners=1ex] at (current bounding box.north
                   west){#1};
        \end{scope}
    }},
}
\tikzset{l3 switch/.style={
    parallelepiped,fill=switch, draw=white,
    minimum width=0.75cm,
    minimum height=0.75cm,
    parallelepiped offset x=1.75mm,
    parallelepiped offset y=1.25mm,
    path picture={
      \node[fill=white,
        circle,
        minimum size=6pt,
        inner sep=0pt,
        append after command={
          \pgfextra{
            \foreach \angle in {0,45,...,360}
            \draw[-latex,fill=white] (\tikzlastnode.\angle)--++(\angle:2.25mm);
          }
        }
      ]
       at ([xshift=-0.75mm,yshift=-0.5mm]path picture bounding box.center){};
    }
  },
  ports/.style={
    line width=0.3pt,
    top color=gray!20,
    bottom color=gray!80
  },
  rack switch/.style={
    parallelepiped,fill=white, draw,
    minimum width=1.25cm,
    minimum height=0.25cm,
    parallelepiped offset x=2mm,
    parallelepiped offset y=1.25mm,
    xscale=-1,
    path picture={
      \draw[top color=gray!5,bottom color=gray!40]
      (path picture bounding box.south west) rectangle
      (path picture bounding box.north east);
      \coordinate (A-west) at ([xshift=-0.2cm]path picture bounding box.west);
      \coordinate (A-center) at ($(path picture bounding box.center)!0!(path
        picture bounding box.south)$);
      \foreach \x in {0.275,0.525,0.775}{
        \draw[ports]([yshift=-0.05cm]$(A-west)!\x!(A-center)$)
          rectangle +(0.1,0.05);
        \draw[ports]([yshift=-0.125cm]$(A-west)!\x!(A-center)$)
          rectangle +(0.1,0.05);
       }
      \coordinate (A-east) at (path picture bounding box.east);
      \foreach \x in {0.085,0.21,0.335,0.455,0.635,0.755,0.875,1}{
        \draw[ports]([yshift=-0.1125cm]$(A-east)!\x!(A-center)$)
          rectangle +(0.05,0.1);
      }
    }
  },
  server/.style={
    parallelepiped,
    fill=white, draw,
    minimum width=0.35cm,
    minimum height=0.75cm,
    parallelepiped offset x=3mm,
    parallelepiped offset y=2mm,
    xscale=-1,
    path picture={
      \draw[top color=gray!5,bottom color=gray!40]
      (path picture bounding box.south west) rectangle
      (path picture bounding box.north east);
      \coordinate (A-center) at ($(path picture bounding box.center)!0!(path
        picture bounding box.south)$);
      \coordinate (A-west) at ([xshift=-0.575cm]path picture bounding box.west);
      \draw[ports]([yshift=0.1cm]$(A-west)!0!(A-center)$)
        rectangle +(0.2,0.065);
      \draw[ports]([yshift=0.01cm]$(A-west)!0.085!(A-center)$)
        rectangle +(0.15,0.05);
      \fill[black]([yshift=-0.35cm]$(A-west)!-0.1!(A-center)$)
        rectangle +(0.235,0.0175);
      \fill[black]([yshift=-0.385cm]$(A-west)!-0.1!(A-center)$)
        rectangle +(0.235,0.0175);
      \fill[black]([yshift=-0.42cm]$(A-west)!-0.1!(A-center)$)
        rectangle +(0.235,0.0175);
    }
  },
}

\usetikzlibrary{calc, shadings, shadows, shapes.arrows}
\tikzset{cross/.style={cross out, draw=black, minimum size=2*(#1-\pgflinewidth), inner sep=0pt, outer sep=0pt},
cross/.default={1pt}}
\tikzset{%
  interface/.style={draw, rectangle, rounded corners, font=\LARGE\sffamily},
  ethernet/.style={interface, fill=yellow!50},% ethernet interface
  serial/.style={interface, fill=green!70},% serial interface
  speed/.style={sloped, anchor=south, font=\large\sffamily},% line speed at edge
  route/.style={draw, shape=single arrow, single arrow head extend=4mm,
    minimum height=1.7cm, minimum width=3mm, white, fill=switch!20,
    drop shadow={opacity=.8, fill=switch}, font=\tiny}% inroute/outroute arrows
}
\newcommand*{\shift}{1.3cm}% For placing the arrows later
\newcommand*{\router}[1]{
\begin{tikzpicture}
  \coordinate (ll) at (-3,0.5);
  \coordinate (lr) at (3,0.5);
  \coordinate (ul) at (-3,2);
  \coordinate (ur) at (3,2);
  \shade [shading angle=90, left color=switch, right color=white] (ll)
    arc (-180:-60:3cm and .75cm) -- +(0,1.5) arc (-60:-180:3cm and .75cm)
    -- cycle;
  \shade [shading angle=270, right color=switch, left color=white!50] (lr)
    arc (0:-60:3cm and .75cm) -- +(0,1.5) arc (-60:0:3cm and .75cm) -- cycle;
  \draw [thick] (ll) arc (-180:0:3cm and .75cm)
    -- (ur) arc (0:-180:3cm and .75cm) -- cycle;
  \draw [thick, shade, upper left=switch, lower left=switch,
    upper right=switch, lower right=white] (ul)
    arc (-180:180:3cm and .75cm);
  \node at (0,0.5){\color{blue!60!black}\Huge #1};% The name of the router
  \begin{scope}[yshift=2cm, yscale=0.28, transform shape]
    \node[route, rotate=45, xshift=\shift] {\strut};
    \node[route, rotate=-45, xshift=-\shift] {\strut};
    \node[route, rotate=-135, xshift=\shift] {\strut};
    \node[route, rotate=135, xshift=-\shift] {\strut};
  \end{scope}
\end{tikzpicture}}

\makeatletter
\pgfdeclareradialshading[tikz@ball]{cloud}{\pgfpoint{-0.275cm}{0.4cm}}{%
  color(0cm)=(tikz@ball!75!white);
  color(0.1cm)=(tikz@ball!85!white);
  color(0.2cm)=(tikz@ball!95!white);
  color(0.7cm)=(tikz@ball!89!black);
  color(1cm)=(tikz@ball!75!black)
}
\tikzoption{cloud color}{\pgfutil@colorlet{tikz@ball}{#1}%
  \def\tikz@shading{cloud}\tikz@addmode{\tikz@mode@shadetrue}}
\makeatother

\tikzset{my cloud/.style={
     cloud, draw, aspect=2,
     cloud color={gray!5!white}
  }
}

\usetikzlibrary{backgrounds}
\usetikzlibrary{patterns}
\pgfmathdeclarefunction{gauss}{3}{%
  \pgfmathparse{1/(#3*sqrt(2*pi))*exp(-((#1-#2)^2)/(2*#3^2))}%
}
\pgfmathdeclarefunction{cdf}{3}{%
  \pgfmathparse{1/(1+exp(-0.07056*((#1-#2)/#3)^3 - 1.5976*(#1-#2)/#3))}%
}
\pgfmathdeclarefunction{fq}{3}{%
  \pgfmathparse{1/(sqrt(2*pi*#1))*exp(-(sqrt(#1)-#2/#3)^2/2)}%
}
\pgfmathdeclarefunction{fq0}{1}{%
  \pgfmathparse{1/(sqrt(2*pi*#1))*exp(-#1/2)}%
}

\usepackage{etoolbox}

\usepackage{wrapfig}

\makeatletter
\newcommand{\setword}[2]{%
  \phantomsection
  #1\def\@currentlabel{\unexpanded{#1}}\label{#2}%
}
\makeatother

\newcommand{\figref}[1]{\hyperref[#1]{Fig.~\ref*{#1}}}
\newcommand{\figsref}[1]{\hyperref[#1]{Figs.~\ref*{#1}}}
\newcommand{\Figref}[1]{\hyperref[#1]{Figure~\ref*{#1}}}
\newcommand{\Figsref}[1]{\hyperref[#1]{Figures~\ref*{#1}}}
\newcommand{\tableref}[1]{\hyperref[#1]{Table~\ref*{#1}}}
\newcommand{\tablesref}[1]{\hyperref[#1]{Tables~\ref*{#1}}}
\newcommand{\appendixref}[1]{\hyperref[#1]{Appendix~\ref*{#1}}}
\newcommand{\theoremref}[1]{\hyperref[#1]{Thm.~\ref*{#1}}}
\newcommand{\Theoremref}[1]{\hyperref[#1]{Theorem~\ref*{#1}}}
\newcommand{\lemmaref}[1]{\hyperref[#1]{Lemma~\ref*{#1}}}
\newcommand{\propref}[1]{\hyperref[#1]{Prop.~\ref*{#1}}}
\newcommand{\propsref}[1]{\hyperref[#1]{Props.~\ref*{#1}}}
\newcommand{\Propsref}[1]{\hyperref[#1]{Propositions~\ref*{#1}}}
\newcommand{\Propref}[1]{\hyperref[#1]{Proposition~\ref*{#1}}}
\newcommand{\corref}[1]{\hyperref[#1]{Cor.~\ref*{#1}}}
\newcommand{\Corref}[1]{\hyperref[#1]{Corollary~\ref*{#1}}}
\newcommand{\scenarioref}[1]{\hyperref[#1]{Scenario~\ref*{#1}}}
\newcommand{\Scenarioref}[1]{\hyperref[#1]{\textsc{scenario}~\ref*{#1}}}
\newcommand{\probref}[1]{\hyperref[#1]{Prob.~\ref*{#1}}}
\newcommand{\Probref}[1]{\hyperref[#1]{Problem~\ref*{#1}}}
\newcommand{\gameref}[1]{\hyperref[#1]{Game~\ref*{#1}}}
\newcommand{\chapterref}[1]{\hyperref[#1]{Chapter~\ref*{#1}}}
\newcommand{\sectionref}[1]{\hyperref[#1]{\S\ref*{#1}}}
\newcommand{\Algref}[1]{\hyperref[#1]{Algorithm ~\ref*{#1}}}
\newcommand{\myalgref}[1]{\hyperref[#1]{Alg.~\ref*{#1}}}
\newcommand{\Myalgref}[1]{\hyperref[#1]{Algorithm~\ref*{#1}}}
\newcommand{\defref}[1]{\hyperref[#1]{Def.~\ref*{#1}}}
\newcommand{\Defref}[1]{\hyperref[#1]{Definition~\ref*{#1}}}
\newcommand{\assumptionref}[1]{\hyperref[#1]{Assumption~\ref*{#1}}}
\newcommand{\remarkref}[1]{\hyperref[#1]{Remark~\ref*{#1}}}
\newcommand{\exampleref}[1]{\hyperref[#1]{Ex.~\ref*{#1}}}
\newcommand{\eqqref}[1]{\hyperref[#1]{Eq.~(\ref*{#1})}}
\newcommand{\eqqqref}[1]{\hyperref[#1]{(\ref*{#1})}}
\newcommand{\eqqsref}[1]{\hyperref[#1]{Eqs.~(\ref*{#1})}}
\newcommand{\Eqqref}[1]{\hyperref[#1]{Equation~(\ref*{#1})}}

\usepackage{arydshln}

\newcommand{\cmark}{\textcolor{OliveGreen}{\ding{51}}}  % Checkmark
\newcommand{\xmark}{\textcolor{Red}{\ding{55}}}    % Cross
\usepackage{lettrine}

\usepackage{fontawesome5}
\usepackage{listofitems} % for \readlist to create arrays
\tikzstyle{mynode}=[thick,draw=blue,fill=blue!20,circle,minimum size=22]

\newtcolorbox{examplebox}{
  colback=black!5!white,
  colframe=black!30!black!70,
  title=Example \textsc{pomdp}: Intrusion recovery,
  fonttitle=\bfseries,
  sharp corners
}

\newtcolorbox{exampleboxtwo}{
  colback=black!5!white,
  colframe=black!30!black!70,
  title=Example feature space,
  fonttitle=\bfseries,
  sharp corners
}

\newtcolorbox{frameworkbox}{
  colback=black!5!white,
  colframe=black!30!black!70,
  title=Framework summary,
  fonttitle=\bfseries,
  sharp corners
}

\usepackage{mdframed}

\definecolor{grad1}{RGB}{0,150,150}
\definecolor{grad2}{RGB}{0,100,200}
\definecolor{grad3}{RGB}{80,0,150}

\usepackage{tikz-3dplot}
\tdplotsetmaincoords{70}{110}

\newtcolorbox{summary}{
  colback=black!5!white,
  colframe=black!30!black!70,
  title=Summary of our method for incident response ({\hypersetup{linkcolor=white}\figref{fig:method}}),
  fonttitle=\bfseries,
  sharp corners
}

\AtBeginDocument{%
  }
\copyrightyear{2025}
\acmYear{2025}
\setcopyright{cc}
\setcctype{by}
\acmConference[AISec '25]{Proceedings of the 2025 Workshop on Artificial Intelligence and Security}{October 13--17, 2025}{Taipei, Taiwan}
\acmBooktitle{Proceedings of the 2025 Workshop on Artificial Intelligence and Security (AISec '25), October 13--17, 2025, Taipei, Taiwan}\acmDOI{10.1145/3733799.3762965}
\acmISBN{979-8-4007-1895-3/2025/10}

\begin{document}
\title{Online Incident Response Planning under Model Misspecification\\through Bayesian Learning and Belief Quantization}

\author{Kim Hammar}
\affiliation{%
  \institution{KTH Royal Institute of Technology}
  \city{Stockholm}
  \country{Sweden}}
\email{kimham@kth.se}

\author{Tao Li}
\affiliation{%d
  \institution{City University of Hong Kong}
  \city{Hong Kong}
  \country{China}
}
\email{li.tao@cityu.edu.hk}

\renewcommand{\shortauthors}{Hammar and Li}

\begin{abstract}
Effective responses to cyberattacks require fast decisions, even when information about the attack is incomplete or inaccurate. However, most decision-support frameworks for incident response rely on a detailed system model that describes the incident, which restricts their practical utility. In this paper, we address this limitation and present an online method for incident response planning under model misspecification, which we call \textsc{mobal}: \underline{M}isspecified \underline{O}nline \underline{Ba}yesian \underline{L}earning. \textsc{mobal} iteratively refines a conjecture about the model through Bayesian learning as new information becomes available, which facilitates model adaptation as the incident unfolds. To determine effective responses online, we quantize the conjectured model into a finite Markov model, which enables efficient response planning through dynamic programming. We prove that Bayesian learning is asymptotically consistent with respect to the information feedback. Additionally, we establish bounds on misspecification and quantization errors. Experiments on the \textsc{cage-2} benchmark show that \textsc{mobal} outperforms the state of the art in terms of adaptability and robustness to model misspecification.
\end{abstract}

\begin{CCSXML}
<ccs2012>
<concept>
<concept_id>10002978.10003014</concept_id>
<concept_desc>Security and privacy~Network security</concept_desc>
<concept_significance>500</concept_significance>
</concept>
</ccs2012>
\end{CCSXML}

\ccsdesc[500]{Security and privacy~Network security}

\keywords{Cybersecurity, reinforcement learning, Bayesian learning, POMDP, misspecification, incident response, network security.}
%\received{20 June 2025}

\maketitle

\section{Introduction}
%\tikzexternaldisable
\begin{figure}
  \centering
%\tikzsetnextfilename{method}
  \scalebox{1}{
    \vspace{0.5cm}
    \input{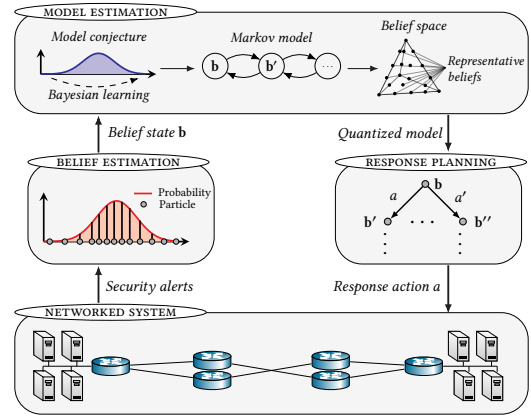}
  }
  \caption{Our method (\textsc{mobal}) for incident response planning under model misspecification. At each time step, we estimate a belief about the system's security state and use it to update a conjecture about the system model through Bayesian learning. We then use this conjecture to sample a specific Markov model, whose belief space is quantized. Finally, we use the quantized model to efficiently compute an effective response through dynamic programming.}
  \label{fig:method}
\end{figure}
%\tikzexternalenable

Incident response refers to the coordinated actions taken to contain, mitigate, and recover from cyberattacks. In practice, incident response is largely a manual process carried out by security experts. Although effective in many cases, this approach is slow, resource-intensive, and requires substantial expertise. For example, a recent study reports that organizations take an average of 73 days to respond and recover from an incident \citep{ibm2024costofdatabreach}. Reducing this delay requires better decision-support systems to assist operators during incident handling. Currently, the standard approach to assisting operators relies on \textit{response playbooks} \citep{10.1145/3491102.3517559}, which comprise predefined rules for handling specific incidents. However, playbooks still rely on security experts for configuration and are therefore difficult to keep aligned with evolving threats and system architectures \citep{10646756}.
%When choosing response actions, a security operator must balance the need to mitigate the attack against the risk of disrupting services. Currently, these decisions are managed through security operations centers, where human operators respond to incidents in real-time with the help of intrusion response systems \citep{wazuh} and incident response playbooks \citep{playbook_response}. While these decision-support systems can be effective, their configuration is labor-intensive and slow.

To address these drawbacks, significant research efforts have started to develop tools for automating the computation of effective incident response strategies for networked systems. This research draws on concepts and methods from various fields, most notably control theory \citep{10955193}, game theory \citep{altman_jamming_1, tao25symbio-model}, dependability \citep{5270307}, large language models (\textsc{llm}s) \citep{10.1145/3689932.3694770,castro2025largelanguagemodelsautonomous, li2025texts,rigaki2023cage}, and reinforcement learning \citep{deep_rl_cyber_sec,beyond_cage,huang2025intentbasedontologydrivenautonomicsecurity}. Broadly speaking, the approach in this line of research is to first construct a model or simulator of the system and then compute an optimal response strategy using numerical methods, such as dynamic programming \citep{tifs_25_HLALB}, reinforcement learning \citep{beyond_cage}, tree search \citep{hammar2024optimaldefenderstrategiescage2}, or \textsc{llm}s \citep{hammar2025incidentresponseplanningusing,castro2025largelanguagemodelsautonomous}. As a consequence, the quality of the resulting response strategy depends critically on the accuracy of the model or simulator, which must capture the system’s (causal) \textit{dynamics}, i.e., how the system evolves in response to attacks and response actions. However, such accurate models and simulators are generally not available in practice due to the complexity of operational systems and the uncertainty about attacks \citep{tao23ztd}. Hence, the practical applicability of current solutions is limited.

In this paper, we address this limitation by presenting an online method for incident response planning under \textit{model misspecification}, which we call \textsc{mobal}: \underline{M}isspecified \underline{O}nline \underline{Ba}yesian \underline{L}earning; see \figref{fig:method}. In particular, we relax the standard assumption that the system model is known and only assume a probabilistic \textit{conjecture} about the model, which may be misspecified in the sense that it assigns $0$ probability to the true model. In our method, this conjecture is iteratively adapted based on available threat information via Bayesian learning. We then use the updated conjecture to compute an effective response strategy using dynamic programming.

A key challenge when performing this computation is the complexity of the dynamic programming problem, which results from two factors: (\textit{i}) the system's security state is only partially observable; and (\textit{ii}) the number of possible system states is large and typically grows exponentially with the system's size. As a consequence, effective incident response requires planning over a high-dimensional \textit{belief space}, i.e., a space of probability distributions over possible states. To address this computational complexity, our method \textit{quantizes} the belief space of the conjectured model, which enables efficient computation of a near-optimal response strategy.

We prove that \textsc{mobal} converges to a conjectured model that is consistent with the observed threat data. Moreover, we derive bounds on both the approximation error (due to quantization) and the misspecification error. To evaluate \textsc{mobal} experimentally, we apply it to \textsc{cage-2} \citep{cage_challenge_2_announcement}, which is a standard benchmark to evaluate incident response frameworks. The results show that \textsc{mobal} offers substantial improvements in adaptability and robustness to model misspecification compared to the state-of-the-art methods.

We summarize our contributions as follows:
\begin{itemize}
\item We develop \textsc{mobal}, an online method for incident response planning under model misspecification. It involves a novel combination of Bayesian learning and belief quantization.
\item We derive theoretical bounds on both the approximation error introduced by the quantization and the error due to model misspecification. We also quantify the interplay between these two errors and establish conditions under which the conjectured model learned by \textsc{mobal} converges.
\item We evaluate \textsc{mobal} on \textsc{cage-2} \citep{cage_challenge_2_announcement}, which involves responding to an advanced persistent threat in an \textsc{it} infrastructure. The results show that \textsc{mobal} outperforms the state-of-the-art in adaptability and robustness to model misspecification.
\end{itemize}

\section{Use Case}\label{sec:use_case}
\begin{figure}[H]
  \centering
%\tikzsetnextfilename{cage_network} 
  \scalebox{1.2}{              
    \input{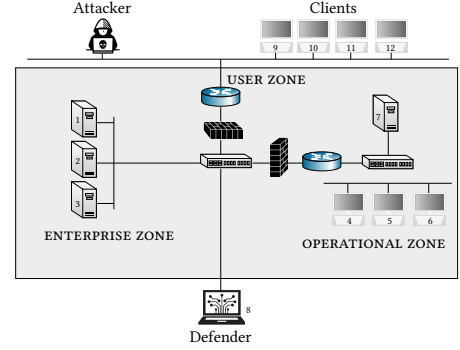}
  }
  \caption{The actors and systems involved in the incident response use case. The system configuration and topology correspond to the \textsc{cage-2} system \citep{cage_challenge_2_announcement}.}
  \label{fig:cage_network}
\end{figure}

We consider a general incident response use case that involves the \textsc{it} infrastructure of an organization. The operator of this infrastructure, which we call the \textit{defender}, takes measures to protect it against an \textit{attacker} while providing services to a client population. An example infrastructure is shown in \figref{fig:cage_network}. This infrastructure is segmented into zones with interconnected servers, which clients access through a public gateway. Though intended for service delivery, this gateway is also accessible to a potential attacker who aims to compromise servers. To achieve this goal, the attacker can perform various actions, such as reconnaissance, brute-force attacks, lateral movement, and exploits (i.e., cyber kill chain \citep{hutchins2011intelligence, tao24ddztd}).

Given these attacker capabilities, we study the problem of developing \textit{optimal incident response strategies} that map infrastructure statistics to automated actions for mitigating potential attacks while minimizing service disruption. Examples of response actions include shutdown, access control, and network resegmentation.

%\tikzexternaldisable
% \tikzexternalenable

%\tikzexternaldisable
\begin{figure*}
  \centering
%\tikzsetnextfilename{pipeline}   
  \scalebox{0.85}{
    \input{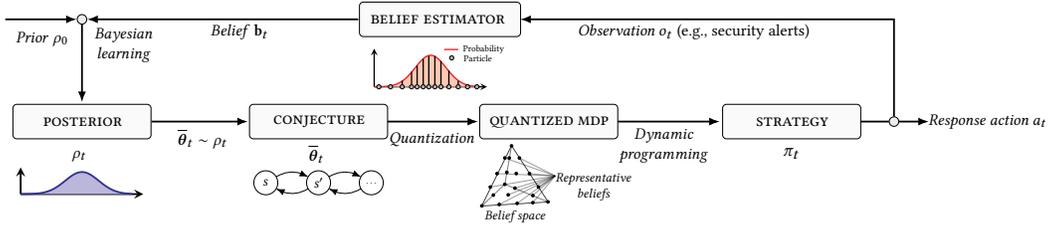}
  }
  \caption{\textsc{mobal}: an iterative method for online learning of incident response strategies under model misspecification. The figure illustrates a time step during which (\textit{i}) the posterior distribution over possible system models is updated via Bayesian learning based on feedback from the system; (\textit{ii}) a conjectured model is sampled from the posterior and quantized into a computationally tractable \textsc{mdp}; and (\textit{iii}) a response strategy is computed using dynamic programming.}
  \label{fig:col}
\end{figure*}

% \tikzexternalenable
\section{Formalizing the Incident Response Use Case}\label{sec:formalization}
We formulate the incident response use case described above as a partially observable Markov decision process (\textsc{pomdp}). Following this formalism, a response strategy $\pi$ is a function that sequentially prescribes \textit{response actions} $a_0,a_1,\hdots$ based on a series of \textit{observations} $o_1,o_2,\hdots$ (e.g., system metrics). These actions stochastically influence the evolution of the system's \textit{state} $s_t$, which captures its security and service status. Due to limited monitoring capabilities or intentional concealment by a potential attacker, the state of the system cannot be observed directly. Therefore, response actions are selected based on a \textit{belief state} $\mathbf{b}_t$, which represents the conditional probability distribution over possible states of the system given observations. The effectiveness of these actions is quantified through a \textit{cost function} that should be minimized.

We denote the set of response actions by $\mathcal{A}$, the set of observations by $\mathcal{O}$, and the set of states by $\mathcal{S}=\{1,\hdots, n\}$, all of which are finite. State transitions $s \rightarrow s^{\prime}$ under action $a$ occur at discrete times $t$ according to transition probabilities $p_{ss^{\prime}}(a)$. Each transition is associated with a real-valued cost $c(s, a)$ and an observation $o$, which is generated with probability $z(o \mid s^{\prime})$. While the \textsc{pomdp} involves imperfect state information, it can be reformulated as an equivalent problem with perfect state information; see e.g., \citep{ASTROM1965174}. In this formulation, the system is described by the belief state $\mathbf{b}=\big(\mathbf{b}(1),\mathbf{b}(2),\dots,\mathbf{b}(n)\big)$, where $\mathbf{b}(i)$ is the conditional probability that the state is $i$, given the history of actions and observations. This vector belongs to the belief space $\mathcal{B}$ and is updated as
\begin{align}
\mathbf{b}_{t}&= \mathbb{B}(\mathbf{b}_{t-1}, a_{t-1}, o_t), \label{eq:belief_estimator}
\end{align}
where $\mathbb{B}$ is a given belief estimator.

We adopt the belief-space formulation and consider response strategies $\pi$ that map the belief space $\mathcal{B}$ to the action space $\mathcal{A}$. Our goal is to minimize the expected discounted cumulative cost, i.e.,
\begin{align}
\minimize_{\pi \in \Pi}\lim_{T \rightarrow \infty}\mathop{\mathbb{E}}_{(s_t,\mathbf{b}_t)_{t\geq 0}}\left\{\sum_{t=0}^{T}\gamma^tc(s_t, \pi(\mathbf{b}_t)) \mid \mathbf{b}_0\right\}, \label{eq:objective}
\end{align}
where $\Pi$ is the strategy space, $\mathbb{E}\{\cdot\}$ denotes the expectation operator, $s_t$ is the state at time $t$, and $\gamma \in (0,1)$ is a discount factor. We say that a strategy $\pi^{\star}$ is optimal if it achieves this minimization. Such a strategy is related to the optimal cost function $J^{\star}$ through the Bellman equations
\begin{subequations}\label{eq:belief_bellman}
\begin{align}
\pi^{\star}(\mathbf{b}) &\in \argmin_{a \in \mathcal{A}}\left[\hat{c}(\mathbf{b}, a) + \gamma \sum_{\mathbf{b}^{\prime} \in \mathcal{B}}p_{\bm{\theta}}(\mathbf{b}^{\prime} \mid \mathbf{b}, a)J^{\star}(\mathbf{b}^{\prime})\right]\label{eq:optimal_strategy},\\
J^{\star}(\mathbf{b}) &= \min_{a \in \mathcal{A}}\left[\hat{c}(\mathbf{b}, a) + \gamma \sum_{\mathbf{b}^{\prime} \in \mathcal{B}}p_{\bm{\theta}}(\mathbf{b}^{\prime} \mid \mathbf{b}, a)J^{\star}(\mathbf{b}^{\prime})\right],\label{eq:optimal_cost}\\
\hat{c}(\mathbf{b}, a)&=\sum_{s=1}^n\mathbf{b}(s)c(s,a), \label{eq:belief_cost}
\end{align}  
\end{subequations}
where $p_{\bm{\theta}}(\mathbf{b}^{\prime} \mid \mathbf{b}, a)$ is the probability of transitioning from belief $\mathbf{b}$ to belief $\mathbf{b}^{\prime}$ when taking action $a$. We assume that the transition probabilities are parameterized by a parameter vector $\bm{\theta}$. Since the transition probabilities depend on the attacker's behavior, we consider the vector $\bm{\theta}$ to be unknown and assume only a probabilistic \textit{conjecture} about $\bm{\theta}$, which we express through a probability distribution $\rho_t$ over some set $\Theta$ of plausible parameter vectors. We say that the conjecture distribution $\rho_t$ is \textit{misspecified} if $\bm{\theta} \not\in \Theta$.
\begin{remark}
Since the state, action and observation spaces are assumed finite, it follows that a) an optimal strategy exists; and b) for each belief $\mathbf{b}$ and action $a$, the transition probability $p_{\bm{\theta}}(\mathbf{b}^{\prime} \mid \mathbf{b}, a)$ is non-zero only for a \textit{finite} set of beliefs $\mathbf{b}^{\prime}$; see e.g., \citep[Thms. 7.6.1--7.6.2]{krishnamurthy_2016} for details. Hence, the Bellman equations in (\ref{eq:belief_bellman}) are well-defined.
\end{remark}

\section{\underline{M}isspecified \underline{O}nline \underline{Ba}yesian \underline{L}earning}\label{sec:method}
Building on the preceding problem formulation, we develop an online method for incident response planning that accounts for misspecification, which we call \textsc{mobal}: \underline{M}isspecified \underline{O}nline \underline{Ba}yesian \underline{L}earning. Our method evolves through a sequence of iterative steps $t=0,1,2,\hdots$, as illustrated in \figref{fig:col}. Each step includes three stages. First, we use the observations $o_1,\hdots,o_t$ (e.g., security alerts) to estimate a belief $\mathbf{b}_t$ about the system state through the belief estimator (\ref{eq:belief_estimator}). Second, we use the same observations to update the distribution $\rho_t$ and conjecture the parameter vector $\bm{\theta}$ [cf.~(\ref{eq:belief_bellman})] as $\overline{\bm{\theta}} \sim \rho_t$. Lastly, we use the conjecture $\overline{\bm{\theta}}$ to construct a computationally tractable Markov decision process (\textsc{mdp}) via belief quantization, which allows us to efficiently approximate an optimal incident response strategy through dynamic programming. These three stages are formally defined next, starting with belief estimation.

\subsection{Belief Estimation}
\label{subsec:particle}
In the context of incident response, the belief state represents a probabilistic estimate of the system's security state, which encodes information about services and possible attacks. Consequently, accurate belief estimation is key to making informed response decisions amidst uncertainty about potential attacks.

The belief state at time $t$ can be computed via the recursion
\begin{align}
\mathbf{b}_t(s^{\prime})&= \frac{z(o_{t} \mid s^{\prime})\sum^{n}_{s=1}\mathbf{b}_{t-1}(s)p_{ss^{\prime}}(a_{t-1})}{\sum_{i=1}^{n}\sum^{n}_{j=1}z(o_{t} \mid j)\mathbf{b}_{t-1}(i)p_{ij}(a_{t-1})}, && \text{for all }s^{\prime}\in \mathcal{S}. \label{eq:bayes_filter}
\end{align}
However, the complexity of this calculation is quadratic in the number of states $n$, which can become computationally intractable for systems with large state spaces. In such cases, the belief state can be efficiently estimated through \textit{particle filtering} as
\begin{align}
\widehat{\mathbf{b}}_t(s) = \frac{1}{M}\sum_{j=1}^M\delta_{s\widehat{s}_t^{(j)}}, && \text{for all }s \in \mathcal{S},\label{eq:estimate_belief}
\end{align}
where $\delta_{ij}=1$ if $i=j$, $\delta_{ij}=0$ if $i\neq j$, and $\widehat{s}_t^{1},\hdots,\widehat{s}_t^{M}$ are states (particles) sampled with probability proportional to the numerator in (\ref{eq:bayes_filter}). Such sampling ensures that the estimated belief $\widehat{\mathbf{b}}_t$ converges (almost surely) to $\mathbf{b}_t$ when $M \rightarrow \infty$. Hence, the particle filter provides a consistent way to estimate beliefs while allowing computational complexity to be adjusted by tuning the number of particles $M$.

\subsection{Bayesian Learning of the System Model}\label{sec:bayesian_learning}
Given the updated belief state, the second step of \textsc{mobal} is to refine the conjecture about the system model based on the observation $o_t$. Specifically, we update the conjecture distribution $\rho_t$ (treated as the probability density function) according to
\begin{align}
\rho_{t}(\overline{\bm{\theta}})&= \frac{P(o_t \mid \overline{\bm{\theta}}, \mathbf{b}_{t-1}, a_{t-1})\rho_{t-1}(\overline{\bm{\theta}})}{\int_{\Theta}P(o_t \mid \bm{\theta}', \mathbf{b}_{t-1},a_{t-1})\rho_{t-1}(\bm{\theta}')\mathrm{d}\bm{\theta}'} && \text{for all } \overline{\bm{\theta}} \in \Theta, \label{eq:bayesian_learning}
\end{align}
where $P(o_t \mid \overline{\bm{\theta}}, \mathbf{b}_{t-1}, a_{t-1})$ is the probability of the observation $o_t$ conditioned on the conjectured parameter vector $\overline{\bm{\theta}}$, the belief state $\mathbf{b}_{t-1}$, and the response action $a_{t-1}$. The goal when refining the conjecture distribution $\rho_t$ in this way is to concentrate probability density on parameter vectors $\overline{\bm{\theta}} \sim \rho_t$ that are consistent with the observations $o_1,\hdots,o_t$. In other words, we seek to minimize the \textit{discrepancy} between the observation distribution in the (conjectured) model parameterized by $\overline{\bm{\theta}}$ and the true model parameterized by $\bm{\theta}$; cf.~(\ref{eq:belief_bellman}). We define this discrepancy as
\begin{equation}
\begin{split}  
  K(\overline{\bm{\theta}},\nu_t) &= \mathbb{E}_{\mathbf{b}\sim \nu_t}\mathbb{E}_{o}\left\{\ln\left(\frac{P(o \mid \bm{\theta}, \mathbf{b})}{P(o \mid \overline{\bm{\theta}}, \mathbf{b})}\right) \mid  \bm{\theta}, \mathbf{b}, \pi\right\},\label{eq:discrepancy}
\end{split}  
\end{equation}
where $P(o\mid \bm{\theta}, \mathbf{b})$ is obtained by marginalizing $P(o\mid \bm{\theta}, \mathbf{b}, a)$ using the empirical action distribution based on the actions $a_0,a_1,\hdots,a_{t-1}$. Similarly, $\nu_t$ denotes the empirical belief distribution\footnote{We use the standard convention that $-\ln 0 = \infty$ and $0\ln 0 = 0$.}, i.e., 
\begin{align*}
  \nu_t(\mathbf{b}) &= \frac{1}{t}\sum_{\tau=1}^{t}\delta_{\mathbf{b}\mathbf{b}_{\tau}}, && \text{for all }\mathbf{b}\in \mathcal{B}.
\end{align*}  
We say that a conjecture $\overline{\bm{\theta}}$ that minimizes the discrepancy $K$ [cf.~(\ref{eq:discrepancy})] is \textit{consistent} \citep{berk, walker01bayesconsis}. Hence, the set of consistent conjectures at time $t$ is given by
\begin{align}
\Theta^\star(\nu_t) = \argmin_{\overline{\bm{\theta}}\in \Theta}K(\overline{\bm{\theta}}, \nu_t). \label{eq:consistent_conjecture_sets}
\end{align}
A desirable property of the posterior $\rho_t$ [cf.~(\ref{eq:bayesian_learning})] is that it concentrates on the consistent conjectures $\Theta^\star(\nu_t)$. This property is guaranteed asymptotically under suitable conditions, as stated below.
\begin{proposition}[Consistent conjectures]\label{label:prop_consistency}
Under suitable regularity conditions (see \appendixref{appendix:regularity}), the following holds
\begin{align}
\label{eq:consistent-conjecture}
\lim_{t\rightarrow\infty}\int_{\Theta} \left(K(\overline{\bm{\theta}},\nu_{t})-K^\star_{\Theta}(\nu_{t})\right)\rho_{t+1}(\overline{\bm{\theta}})\mathrm{d}\overline{\bm{\theta}}&=0 \text{ almost surely},
\end{align}
where $K^\star_{\Theta}(\nu_{t})$ is a finite constant defined as
\begin{align*}
K^\star_{\Theta}(\nu_{t}) = \min_{\overline{\bm{\theta}}\in \Theta}K(\overline{\bm{\theta}}, \nu_{t}).
\end{align*} 
\end{proposition}
While \propref{label:prop_consistency} ensures that the posterior $\rho_t$ eventually concentrates on consistent conjectures [cf.~(\ref{eq:consistent_conjecture_sets})], it does not quantify how close the dynamics induced by these conjectures are to the true system dynamics. In particular, if the true parameter vector $\bm{\theta}$ lies outside the set $\Theta$, then even the most consistent conjecture may yield a transition model $p_{\overline{\bm{\theta}}}(\mathbf{b}' \mid \mathbf{b}, a)$ that deviates significantly from the true model $p_{\bm{\theta}}(\mathbf{b}' \mid \mathbf{b}, a)$. As a result, a response strategy derived from such a conjecture may be suboptimal. To formalize this suboptimality, let $\overline{J}^{\star}$ denote the optimal cost function in the model defined by $\overline{\bm{\theta}}$; cf. (\ref{eq:optimal_cost}). We refer to the difference between this cost function and the optimal cost function $J^{\star}$ [cf.~(\ref{eq:optimal_cost})] as the \textit{misspecification error}. This error is bounded by the difference between $p_{\overline{\bm{\theta}}}$ and $p_{\bm{\theta}}$, as stated in the following proposition.
\begin{proposition}[Misspecification error bound]\label{prop:misspecification_bound}
If the transition probability distributions $p_{\bm{\theta}}$ and $p_{\overline{\bm{\theta}}}$ satisfy
\begin{align}
\sum_{\mathbf{b}^{\prime} \in \mathcal{B}} |p_{\bm{\theta}}(\mathbf{b}^{\prime} \mid \mathbf{b}, a) - p_{\overline{\bm{\theta}}}(\mathbf{b}^{\prime} \mid \mathbf{b}, a)| \leq \alpha,\text{ for all }\mathbf{b} \in \mathcal{B},a\in \mathcal{A},\label{eq:alpha_def}
\end{align}
for some constant $\alpha \in [0,2]$. Then we have
\begin{align*}
\norm{\overline{J}^{\star} - J^{\star}}_{\infty} \leq \frac{\gamma \alpha c_{\textsc{max}}}{(1-\gamma)^2},
\end{align*}
where $\gamma$ is the discount factor and $c_{\textsc{max}}$ is a finite constant defined by
\begin{align}
c_{\textsc{max}} = \max_{\mathbf{b} \in \mathcal{B},a \in \mathcal{A}}\hat{c}(\mathbf{b},a).\label{eq:c_max_def}
\end{align}
\end{proposition}
This proposition quantifies the cost of relying on a misspecified model. It states that the misspecification error grows proportionally with the error of the conjectured state transitions; cf.~(\ref{eq:alpha_def}).
\subsection{Model Quantization and Response Planning}\label{sec:quantization}
Given the updated belief and conjecture, the last step of \textsc{mobal} is to compute an effective response strategy. While an optimal strategy (according to the conjecture $\overline{\bm{\theta}}$) can (in principle) be computed using dynamic programming techniques, this computation is intractable due to the continuous belief space $\mathcal{B}$. To circumvent this intractability, we \textit{quantize} $\mathcal{B}$ into a finite set of representative beliefs. Specifically, we define the set of representative beliefs as
\begin{align}
&\Tilde{\mathcal{B}}=\left\{\tilde{\mathbf{b}} \text{ }\bigg|\text{ } \tilde{\mathbf{b}} \in \mathcal{B}, \tilde{\mathbf{b}}(s)=\frac{\beta_{s}}{r}, \sum_{s\in \mathcal{S}}\beta_s = r,\beta_s\in \{0,\hdots,r\}\right\}, \label{eq:aggregate_belief_space}
\end{align}
where $r \in \{1,2,\hdots\}$ is a given parameter that can be interpreted as the \textit{quantization resolution}. To relate the beliefs $\mathbf{b} \in \mathcal{B}$ to the representative beliefs $\tilde{\mathbf{b}} \in \Tilde{\mathcal{B}}$, we define a mapping $\Phi: \mathcal{B} \mapsto \Tilde{\mathcal{B}}$ as
\begin{align}
\Phi(\mathbf{b}) = \argmin_{\tilde{\mathbf{b}} \in \Tilde{\mathcal{B}}}\norm{\mathbf{b}-\tilde{\mathbf{b}}}_{\infty}, && \text{for all }\mathbf{b}\in \mathcal{B},\label{eq:aggregation_mapping}
\end{align}
where ties in the argmin are broken in a consistent way. Given this mapping from the belief space $\mathcal{B}$ to the set of representative beliefs $\Tilde{\mathcal{B}}$, we obtain a well-defined \textsc{mdp} whose state space is the set of representative beliefs $\Tilde{\mathcal{B}}$. The cost function in this \textsc{mdp} is given by (\ref{eq:belief_cost}) and the transition probabilities are defined as
\begin{align*}
\hat{p}_{\overline{\bm{\theta}}}(\tilde{\mathbf{b}}^{\prime} \mid \tilde{\mathbf{b}}, a) = \sum_{\mathbf{b}^{\prime}\in \mathcal{B}}p_{\overline{\bm{\theta}}}(\mathbf{b}^{\prime} \mid \tilde{\mathbf{b}}, a)\delta_{\tilde{\mathbf{b}^{\prime}}\Phi(\mathbf{b}^{\prime})}, && \text{for all }a\in\mathcal{A},\tilde{\mathbf{b}}^{\prime},\tilde{\mathbf{b}}\in \Tilde{\mathcal{B}}.
\end{align*}  
Due to the finite state space, the quantized \textsc{mdp} can be efficiently solved using dynamic programming. Let $V^{\star}$ and $\mu^{\star}$ denote the optimal cost function and strategy in this \textsc{mdp}, respectively. Similarly, let $\overline{J}^{\star}$ and $\overline{\mu}^{\star}$ denote the optimal cost function and strategy of the (non-quantized) \textsc{pomdp} based on the conjecture $\overline{\bm{\theta}}$, respectively; cf.~(\ref{eq:belief_bellman}). We can then approximate $\overline{J}^{\star}$ and $\overline{\mu}^{\star}$ as
\begin{align}
\Tilde{J}(\mathbf{b}) &= V^{\star}(\Phi(\mathbf{b})) \text{ and } \tilde{\pi}(\mathbf{b})=\mu^{\star}(\Phi(\mathbf{b})), \text{ for all } \mathbf{b} \in \mathcal{B}.\label{eq:approximation_1}
\end{align}
We refer to the difference between the cost function approximation $\tilde{J}$ and the (conjectured) optimal cost function $\overline{J}^{\star}$ as the \textit{approximation error}. To understand this error, note that the mapping $\Phi$ [cf.~(\ref{eq:aggregation_mapping})] partitions the belief space $\mathcal{B}$ into disjoint subsets as
\begin{align}
\mathcal{B} = \bigcup_{\tilde{\mathbf{b}} \in \tilde{\mathcal{B}}}S_{\tilde{\mathbf{b}}}, && \text{where } S_{\tilde{\mathbf{b}}}=\left\{\mathbf{b} \mid \mathbf{b} \in \mathcal{B}, \Phi(\mathbf{b}) = \tilde{\mathbf{b}}\right\}.\label{eq:partitioning}
\end{align}
In view of (\ref{eq:approximation_1}), this partitioning means that the approximation error of $\Tilde{J}$ is determined by how much the (conjetured) optimal cost function $\overline{J}^{\star}(\mathbf{b})$ varies for beliefs $\mathbf{b}$ within the same belief-space partition $S_{\tilde{\mathbf{b}}}$. This insight is formalized by the following proposition.
\begin{proposition}[Approximation error bound]\label{prop:aggregation_bound}
The error of the cost function approximation $\Tilde{J}$ [cf.~(\ref{eq:approximation_1})] with respect to the conjectured optimal cost function $\overline{J}^{\star}$ is bounded as
\begin{align*}
|\Tilde{J}(\mathbf{b}) - \overline{J}^{\star}(\mathbf{b})| \leq \frac{\epsilon}{1-\gamma}, && \text{for all } \mathbf{b} \in S_{\tilde{\mathbf{b}}}, \tilde{\mathbf{b}} \in \Tilde{\mathcal{B}},
\end{align*}
where $\gamma$ is the discount factor and $\epsilon$ is a finite constant defined by
\begin{align}
\epsilon = \max_{\tilde{\mathbf{b}} \in \Tilde{\mathcal{B}}}\sup_{\mathbf{b},\mathbf{b}^{\prime} \in S_{\tilde{\mathbf{b}}}}|\overline{J}^{\star}(\mathbf{b})-\overline{J}^{\star}(\mathbf{b}^{\prime})|.\label{eq:eps_def}
\end{align}
\end{proposition}
The meaning of \propref{prop:aggregation_bound} is that the error of the cost function approximation $\tilde{J}$ [cf.~(\ref{eq:approximation_1})] is small if the aggregation mapping $\Phi$ [cf.~(\ref{eq:aggregation_mapping})] conforms to the (conjectured) optimal cost function $\overline{J}^{\star}$ in the sense that $\Phi$ varies little in regions of the belief space where $\overline{J}^{\star}$ also varies little. This error can be controlled by tuning the quantization resolution $r$ [cf.~(\ref{eq:aggregate_belief_space})], as stated in the following proposition.
\begin{proposition}[Asymptotic (conjectured) optimality]\label{prop:consistent}
Given the cost function approximation $\Tilde{J}$ [cf.~(\ref{eq:approximation_1})], the following holds.
\begin{align*}
&\lim_{r \rightarrow \infty} |\Tilde{J}(\mathbf{b}) - \overline{J}^{\star}(\mathbf{b})| = 0, && \text{for all }\mathbf{b} \in \mathcal{B}.
\end{align*}
\end{proposition}
While the preceding propositions quantify the difference between the cost function approximation $\Tilde{J}$ [cf.~(\ref{eq:approximation_1})] and the \textit{conjectured} optimal cost function $\overline{J}^{\star}$, they do not say anything about the difference to the optimal cost function $J^{\star}$. This difference depends on both the approximation error $\norm{\overline{J}^{\star}-\Tilde{J}}_{\infty}$ and the misspecification error $\norm{\overline{J}^{\star}-J^{\star}}_{\infty}$, as captured by the following theorem. 
\begin{theorem}[Sub-optimality bound of \textsc{mobal}]
The sub-optimality of the cost function approximation $\Tilde{J}$ [cf.~(\ref{eq:approximation_1})] is bounded as 
\begin{align*}
\norm{\Tilde{J}-J^{\star}}_{\infty} \leq \frac{\epsilon}{1-\gamma} + \frac{\gamma \alpha c_{\textsc{max}}}{(1-\gamma)^2},
\end{align*}
where $\gamma$ is the discount factor and $(\epsilon,\alpha,c_{\textsc{max}})$ are the finite constants defined in (\ref{eq:eps_def}), (\ref{eq:alpha_def}), and (\ref{eq:c_max_def}), respectively.
\end{theorem}
\begin{proof}
By \propref{prop:aggregation_bound}, we have
\begin{align*}
\norm{\Tilde{J} - \overline{J}^{\star}}_{\infty} \leq \frac{\epsilon}{1-\gamma},
\end{align*}  
and by \propref{prop:misspecification_bound}, we have
\begin{align*}
\norm{\overline{J}^{\star} - J^{\star}}_{\infty} \leq \frac{\gamma \alpha c_{\textsc{max}}}{(1-\gamma)^2}.
\end{align*}
Applying the triangle inequality, we obtain
\begin{align*}
\norm{\Tilde{J} - J^{\star}}_{\infty}\leq \norm{\Tilde{J} - \overline{J}^{\star}}_{\infty} + \norm{\overline{J}^{\star} - J^{\star}}_{\infty} \leq \frac{\epsilon}{1-\gamma} + \frac{\gamma \alpha c_{\textsc{max}}}{(1-\gamma)^2}.
\end{align*}
\end{proof}
This theorem shows that the sub-optimality of \textsc{mobal} decomposes into two components: one due to model approximation ($\epsilon$) and one due to model misspecification ($\alpha$). It is significant because it shows that performance guarantees can be obtained even when relaxing the standard assumption of a correctly specified model.

%\tikzexternaldisable
\begin{summary}
\textsc{mobal} starts with an initial conjecture $\rho_0$ about the incident and a belief $\mathbf{b}_0$ about the security state. Given these priors, \textsc{mobal} proceeds through a sequence of iterative steps $t=0,1,2,\hdots$, where each step consists of three stages. 
\begin{enumerate}
\item The belief $\mathbf{b}_t$ is updated based on the latest observation $o_t$ through recursive state estimation; cf. (\ref{eq:estimate_belief}).
\item The conjecture distribution $\rho_t$ is adapted to the observation $o_t$ through Bayesian learning; cf.~(\ref{eq:bayesian_learning}).
\item A conjecture is sampled $\overline{\bm{\theta}}_t \sim \rho_t$ and used to approximate an optimal response action $a_t$ through dynamic programming and belief quantization; cf.~(\ref{eq:approximation_1}).
\end{enumerate}
\end{summary}
%\tikzexternalenable

\section{Illustrative Example}\label{sec:example}
\begin{figure}[H]
  \centering
%\tikzsetnextfilename{example}   
  \scalebox{0.8}{             
    \input{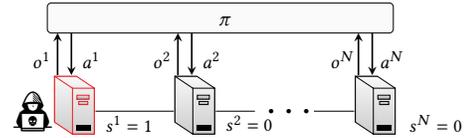}
  }
  \caption{The system in the illustrative example \citep{tifs_25_HLALB}.}
  \label{fig:example}
\end{figure}

To illustrate our method for incident response planning under model misspecification, we consider the response scenario introduced in \citep{tifs_25_HLALB}. This scenario consists of a networked system with $N$ components; see \figref{fig:example}. Each component has two states: $1$ (compromised) or $0$ (safe), i.e., $s=(s^1,\hdots,s^N)$ where $s^l \in \{0,1\}$. Compromises occur randomly over time and incur operational costs. Intrusion detection systems generate observations $o = (o^1, \ldots, o^N)$ that provide partial indications of the components' states, where $o^l \in \{0,1,\hdots\}$ is the number of security alerts related to component $l$. The security policy $\pi$ prescribes the action vector $a=(a^1,\hdots,a^N)$, where each $a^l$ determines whether to block network traffic to component $l$ ($a^l = 1$) or take no action ($a^l = 0$). Blocking a component can prevent further compromise or lateral movement by an attacker, but may also disrupt legitimate services. The goal is to determine a response strategy that balances this trade-off optimally.

We capture this objective through the cost function
\begin{align}
c(s, a) = \sum_{l=1}^{N}\overbrace{2s^l(1-a^l)}^{\text{intrusion cost}} + \overbrace{a^l}^{\text{blocking cost}},\label{example_pomdp_cost}
\end{align}
which encodes that costs are incurred for unmitigated intrusions ($s^l=1$) and for blocking network traffic ($a^l=1$).

%\tikzexternaldisable
% \tikzexternalenable

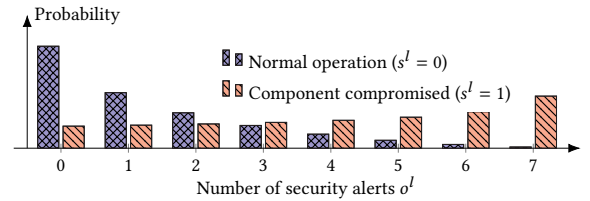
\begin{figure}[H]
  \centering
%\tikzsetnextfilename{obs_dist}   
  \scalebox{0.8}{
    \begin{tikzpicture}
\begin{axis}[
    ybar,
    bar width=10pt,
    ymin=0,
    ymax=0.55,
    xtick={0,1,...,7},
    xticklabels={0,1,2,3,4,5,6,7},    
    xtick=data,
    axis y line=center,
    axis x line=bottom,
    ytick=\empty,
    ylabel={},
    axis line style={-{Latex[length=2mm]}},
    width=11cm,
    height=3.8cm,
    enlarge x limits=0.1,
    legend style={/tikz/every node/.style={anchor=west}, at={(0.9, 0.8)}, legend columns=1, draw=none, /tikz/column 2/.style={
                column sep=10pt,
              }},
]
\addplot+[Black,fill=Blue!40, postaction={
        pattern=crosshatch
      }] coordinates {
(0.5, 0.4204381193405085)
(1.5, 0.22890519830761025)
(2.5, 0.14592706392110147)
(3.5, 0.09381025537785098)
(4.5, 0.05784965748300809)
(5.5, 0.03262720682041655)
(6.5, 0.015497923239697894)
(7.5, 0.0049445755098083705)
  };

\addplot+[Black,fill=Red!40, postaction={
        pattern=north west lines
      }] coordinates {
(0.5, 0.0909090909090911)
(1.5, 0.09497964721845345)
(2.5, 0.09997857601942456)
(3.5, 0.10636018725470703)
(4.5, 0.1149839862213048)
(5.5, 0.12775998469033872)
(6.5, 0.15030586434157509)
(7.5, 0.21472266334510712)
  };  
  \legend{Normal operation ($s^l=0$), Component compromised ($s^l=1$)}
\end{axis}
\node[inner sep=0pt,align=center, scale=1, rotate=0, opacity=1] (obs) at (4.9,-0.65)
{
  Number of security alerts $o^l$
};
\node[inner sep=0pt,align=center, scale=1, rotate=0, opacity=1] (obs) at (1.1,2.2)
{
  Probability
};
\end{tikzpicture}
  }
  \caption{Observation distribution per component $l$ in the illustrative example.}
  \label{fig:obs_dist}
\end{figure}

We define the observation distribution for each component using the Beta-binomial distribution shown in \figref{fig:obs_dist}. Specifically, we define the distribution of $o^l$ as $\mathrm{BetaBin}(7,1,0.7)$ when $s^l=1$ and define the distribution as $\mathrm{BetaBin}(7,0.7,3)$ when $s^l=0$. These distributions reflect that alerts may occur ($o^l > 0$) during normal operation ($s^l=0$) but are more likely during attacks ($s^l=1$). Similar alert distributions have been observed in practice; see e.g., \citep{dsn24_hammar_stadler}.

%\tikzexternaldisable
% \tikzexternalenable

The transition probabilities $p_{ss^{\prime}}(a)$ are defined as follows. If component $l$ is compromised ($s^l=1$), then it remains so until recovery is applied ($a^l=1$), at which point the state $s^l$ is set to $0$. Otherwise, the probability that it becomes compromised is $\min\{p_{\mathrm{A}}(1+\mathcal{N}_l(s)), 1\}$, where $\mathcal{N}_l(s)$ is the number of compromised neighbors of component $l$ in the network and $p_{\mathrm{A}} \in (0,1]$ is a given parameter. This compromise probability reflects how attacks can propagate through neighboring components in the network.

For the numerical examples presented in the following, we consider the case where all parameters of the model are known except $p_{\mathrm{A}}$, which we define as $p_{\mathrm{A}}=0.2$. We define the initial conjecture distribution of this parameter to be a uniform distribution over the set $\Theta = \{0, 0.5, 1\}$, i.e., $\rho_0(0)=\rho_0(0.5)=\rho_0(1)=\frac{1}{3}$;  cf.~(\ref{eq:bayesian_learning}). 

\subsubsection*{\textbf{Numerical examples}} 
\begin{figure}[H]
  \centering
%\tikzsetnextfilename{particle_eval}  
  \scalebox{0.77}{
    \begin{tikzpicture}

\pgfplotsset{/dummy/workaround/.style={/pgfplots/axis on top}}

\pgfplotstableread{
0 0.937409619572725 0.9583945886087313 0.9164246505367186
1 0.21401563349159472 0.28023888895396054 0.14779237802922893
2 0.11660489806477921 0.17216222868003395 0.06104756744952446
3 0.09794059954739116 0.1235502767441885 0.07233092235059382
4 0.1037991331429636 0.1559379884155731 0.05166027787035411
5 0.08360510061788937 0.10859933403274898 0.05861086720302976
6 0.06398805808091008 0.08099305821434702 0.04698305794747315
7 0.06684861153121538 0.0893544518905174 0.04434277117191336
8 0.08889753976183626 0.11969730213109825 0.05809777739257427
9 0.08537910864805243 0.10739437140129389 0.06336384589481096
10 0.06669064584116821 0.08870129447930616 0.04467999720303026
11 0.06512538726361453 0.08252639104627554 0.04772438348095351
12 0.0578512480905915 0.06896789496796504 0.04673460121321796
13 0.042880392474990044 0.051091364153393035 0.03466942079658705
14 0.05339014054870957 0.06560962344526511 0.04117065765215404
15 0.05218378908560498 0.07104950378982447 0.033318074381385496
16 0.05303410360272544 0.07838523444299521 0.027682972762455665
17 0.046746938671509065 0.0557420182944131 0.03775185904860503
18 0.05055725571911864 0.07387682244920013 0.02723768898903715
}\particleonestate

\pgfplotstableread{
0 0.8723677421168713 0.8942851729015479 0.8504503113321947
1 0.32970463785126886 0.40288139307621107 0.25652788262632664
2 0.2547102349416701 0.3086559617516401 0.2007645081317001
3 0.19110871640568014 0.22561815559551524 0.15659927721584505
4 0.15680595509664055 0.1939658123603941 0.11964609783288702
5 0.1770283926840815 0.20770403442042507 0.14635275094773792
6 0.12900423708455072 0.16498395411795194 0.0930245200511495
7 0.11785312517132003 0.15143081638548975 0.08427543395715031
8 0.12186734945987591 0.14991969591831553 0.0938150030014363
9 0.1126977945679392 0.13668150506485297 0.08871408407102543
10 0.11136441131489608 0.1269482427582229 0.09578057987156928
11 0.1250789163443869 0.15515782066504555 0.09500001202372824
12 0.102107423888676 0.1341197732857354 0.07009507449161659
13 0.09312463428047882 0.11944207688690042 0.06680719167405721
14 0.07933184185178899 0.0972590472116541 0.061404636491923875
15 0.08487458323785584 0.10110296017436016 0.06864620630135151
16 0.08016760188772905 0.1070339169729485 0.05330128680250959
17 0.09883340868987589 0.12314543804354676 0.07452137933620502
18 0.08789760165370451 0.11300560450483285 0.06278959880257617
}\particletwostate

\pgfplotstableread{
0 0.8249775979048802 0.8602775814804452 0.7896776143293153
1 0.38892236234448335 0.5046894847402212 0.27315523994874547
2 0.2736654457096794 0.34608559578617126 0.2012452956331875
3 0.2521793873517028 0.3156032478106908 0.18875552689271483
4 0.2313083909826043 0.2583981864568667 0.20421859550834193
5 0.2031175060529072 0.24109303338990157 0.16514197871591285
6 0.17340486525586582 0.1946807307606678 0.15212899975106384
7 0.1822360281530354 0.21824001372301352 0.1462320425830573
8 0.14959471351714085 0.18039842139870538 0.1187910056355763
9 0.17342740227669756 0.2001580818765367 0.14669672267685843
10 0.15316410345979287 0.18248282818803155 0.12384537873155418
11 0.1518560089134428 0.181139967089789 0.12257205073709662
12 0.1326136931056434 0.15748209065672933 0.10774529555455745
13 0.11310643034804493 0.13090907037817268 0.09530379031791719
14 0.12793629890804364 0.1474531431295673 0.10841945468651999
15 0.11217890739719574 0.14021275142385448 0.08414506337053701
16 0.12000301307867271 0.1448691291187426 0.0951368970386028
17 0.10443772154955419 0.12172954004821115 0.08714590305089723
18 0.10027262403573886 0.11926573187316955 0.08127951619830817
}\particlethreestate

\node[scale=1] (kth_cr) at (0,0)
{
\begin{tikzpicture}
  \begin{axis}
[
        xmin=0,
        xmax=19,
        ymax=1.1,
        %ymode=log,
        width=11.5cm,
        height=3.5cm,
        axis y line=center,
        axis x line=bottom,
        scaled y ticks=false,
        yticklabel style={
        /pgf/number format/fixed,
        /pgf/number format/precision=5
      },
        xlabel style={below right},
        ylabel style={above left},
        axis line style={-{Latex[length=2mm]}},
        smooth,
        legend style={at={(0.85,0.7)}},
        legend columns=3,
        legend style={
          draw=none,
            /tikz/column 2/.style={
                column sep=5pt,
              }
              }
              ]
              \addplot[RoyalAzure,name path=l1, thick] table [x index=0, y index=1, domain=0:1] {\particleonestate};
              \addplot[OliveGreen,name path=l1, thick] table [x index=0, y index=1, domain=0:1] {\particletwostate};
              \addplot[Red,name path=l1, thick] table [x index=0, y index=1, domain=0:1] {\particlethreestate};
              \addplot[draw=none,Black,mark repeat=2, name path=A, thick, domain=0:1] table [x index=0, y index=2] {\particleonestate};
              \addplot[draw=none,Black,mark repeat=2, name path=B, thick, domain=0:1] table [x index=0, y index=3] {\particleonestate};
              \addplot[RoyalAzure!10!white] fill between [of=A and B];

              \addplot[draw=none,Black,mark repeat=2, name path=A, thick, domain=0:1] table [x index=0, y index=2] {\particletwostate};
              \addplot[draw=none,Black,mark repeat=2, name path=B, thick, domain=0:1] table [x index=0, y index=3] {\particletwostate};
              \addplot[OliveGreen!10!white] fill between [of=A and B];

              \addplot[draw=none,Black,mark repeat=2, name path=A, thick, domain=0:1] table [x index=0, y index=2] {\particlethreestate};
              \addplot[draw=none,Black,mark repeat=2, name path=B, thick, domain=0:1] table [x index=0, y index=3] {\particlethreestate};
              \addplot[Red!10!white] fill between [of=A and B];

\legend{$n=2$, $n=4\text{ }\text{ }$, $n=8$}
  \end{axis}
\node[inner sep=0pt,align=center, scale=1, rotate=0, opacity=1] (obs) at (1.75,1.95)
{
$\mathbb{E}\left[\norm{\mathbf{b}_t - \widehat{\mathbf{b}}_t}_2\right]$ ($\downarrow$ better)
};
\node[inner sep=0pt,align=center, scale=1, rotate=0, opacity=1] (obs) at (4.75,-0.75)
{
  Number of particles $M$
};
\end{tikzpicture}
};

\end{tikzpicture}
 }
  \caption{Expected error of the particle filter for the illustrative example in function of the number of particles $M$ for varying sizes of the state space $n$ (corresponding to $N=1$, $N=2$, and $N=3$ system components); cf.~(\ref{eq:estimate_belief}). Curves show the mean value from evaluations with $100$ random seeds; shaded areas indicate standard deviations and $\norm{\cdot}_2$ denotes the Euclidean norm. The belief $\mathbf{b}_t$ is calculated using formula (\ref{eq:bayes_filter}) with $\overline{\bm{\theta}} = \bm{\theta}$. We calculate the expectation by running $100$ \textsc{pomdp} episodes of $100$ time steps each with strategy $\tilde{\pi}$ [cf.~(\ref{eq:approximation_1})] computed using quantization resolution $r=5$.}
  \label{fig:particle_eval}
\end{figure}
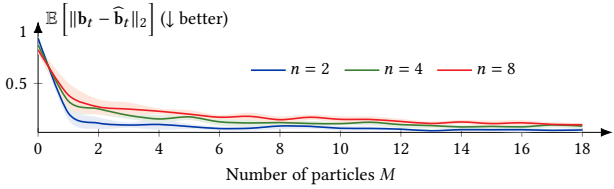

We start by evaluating the accuracy of the particle filter (\ref{eq:estimate_belief}). \Figref{fig:particle_eval} shows the difference between the estimated belief and the true belief. As expected, the accuracy improves with the number of particles $M$. For small systems (e.g., $N=3$ components), we find that the particle filter provides a close approximation to the true belief with only $M=10$ particles.
%\tikzexternaldisable
% \tikzexternalenable

\begin{figure}
  \centering
%\tikzsetnextfilename{bayes_eval_1}  
  \scalebox{0.77}{
    \input{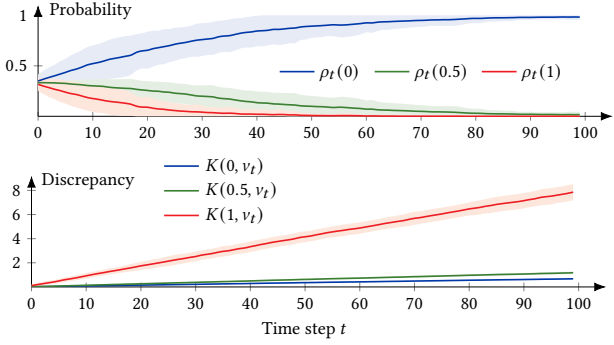}
  }
  \caption{Evolution of the posterior conjecture distribution $\rho_t$ [cf.~(\ref{eq:bayesian_learning})] and the discrepancy $K(\overline{\bm{\theta}},\nu_t)$ [cf.~(\ref{eq:discrepancy})] for the illustrative example. In this example, the true parameter vector is $\bm{\theta}=0.2$, the set of conjectures is $\Theta=\{0, 0.5, 1\}$, and the quantization resolution is $r=5$.}
  \label{fig:bayes_eval}
\end{figure}

Now consider the Bayesian learning formula (\ref{eq:bayesian_learning}). \Figref{fig:bayes_eval} shows the evolution of the posterior $\rho_t$ [cf.~(\ref{eq:bayesian_learning})] and the discrepancy $K$ [cf.~(\ref{eq:discrepancy})]. We observe that the posterior $\rho_t$ converges to a distribution that concentrates on the conjecture $\overline{\bm{\theta}}=0$, which is the conjecture with the lowest discrepancy, as expected from \propref{label:prop_consistency}.
%\tikzexternaldisable
% \tikzexternalenable

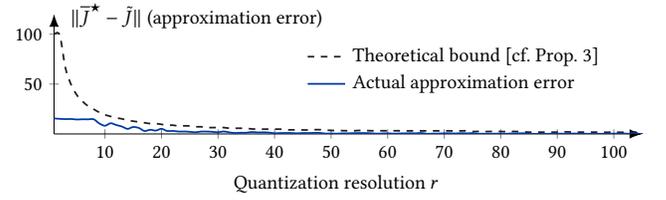
\begin{figure}
  \centering
%\tikzsetnextfilename{rep21}    
  \scalebox{0.83}{
    \begin{tikzpicture}

\pgfplotstableread{
1 99.99999999999991 15.59032754707749
2 99.2 15.131978257743441
3 66.40000000000009 14.878149381086327
4 49.599999999999824 14.987060306964262
5 39.20000000000009 14.498233460516527
6 32.80000000000027 14.70076842778196
7 28.00000000000009 14.657026545439903
8 24.799999999999912 14.73899051984282
9 21.60000000000009 10.533810113630164
10 19.200000000000355 8.219827741472114
11 17.600000000000357 10.728272365688564
12 16.000000000000178 8.964688431215862
13 15.200000000000088 7.446058298316169
14 13.600000000000444 5.094839370515018
15 12.800000000000178 7.013823736191245
16 12.000000000000266 5.99527585134838
17 11.200000000000355 2.9475991171371874
18 10.400000000000444 4.181468576298
19 10.399999999999912 3.394966258060059
20 9.600000000000355 5.170459281044618
21 8.800000000000267 2.9488590156075887
22 8.800000000000267 3.0741377100116267
23 8.000000000000533 2.421367248432425
24 8.139183462754566 2.462223405937806
25 8.0 2.1548714186169207
26 7.200000000000444 1.7048844577456457
27 7.200000000000267 2.4017490821188527
28 6.400000000000356 2.4132089673194272
29 6.400000000000533 2.1457036606504314
30 6.400000000000356 1.7069564146969807
31 6.399999999999823 2.5481752474610175
32 5.717158898083413 1.5573232372447592
33 5.717158898083413 0.9668791245237678
34 5.717158898083413 1.223935063034876
35 5.6000000000000885 1.2393117396258653
36 4.994343713374678 1.9506808463542278
37 4.855548279298456 1.5212401454694877
38 4.800000000000355 1.6287327962923346
39 4.839553308995543 1.3855034887966635
40 4.8000000000001775 0.7453204724509295
41 4.8 0.9102934836873473
42 4.127786596361457 0.9889761791353227
43 4.127786596361457 0.9397184986737717
44 4.000000000000444 0.6485949986162787
45 4.0000000000002665 0.9692215802081883
46 4.0000000000002665 0.9095333920205562
47 4.280942581266341 1.1073576481560838
48 4.280942581266341 1.0641370079217936
49 4.0000000000002665 0.41350580908560985
50 3.484061714578243 0.4660114845968799
51 3.484061714578243 0.3144516625920488
52 3.3294013165507477 0.623688767071032
53 3.200000000000533 0.606890682372029
54 3.200000000000533 0.8482398156819251
55 3.509690318970901 0.9071086615244255
56 3.509690318970901 0.9139511735246106
57 3.200000000000533 0.705991934594012
58 3.2000000000003554 0.35590197605812435
59 3.2000000000003554 0.4779053747679676
60 3.200000000000178 0.6681266096253271
61 3.2 0.31412348327215867
62 3.2 0.4661298086756638
63 3.1155644298966196 0.6123980843925771
64 3.1155644298966196 0.7263208339706022
65 3.1155644298966196 0.6590344785725275
66 3.1155644298966196 1.0117981086298133
67 3.1155644298966196 0.14417136183732993
68 3.1155644298966196 0.3690584782405484
69 3.1155644298966196 0.392340501786947
70 3.1155644298966196 0.6812810745040636
71 3.1155644298966196 0.6841710556976395
72 3.1155644298966196 0.4850777707318148
73 3.1155644298966196 0.8152449280632368
74 3.1155644298966196 0.579229965223437
75 2.4000000000002664 0.8486623387421925
76 2.535418626203343 0.20017186801266362
77 2.4000000000002664 0.26360307342160283
78 2.4575387997581166 0.17540709903825302
79 2.6510512128378827 0.3050806695172401
80 2.4000000000000887 0.3103802853457722
81 2.4000000000000887 0.2299169756295285
82 2.5273270197672537 0.5549180546179482
83 2.4000000000000887 0.5384055754339307
84 2.053480616338808 0.7288207133024702
85 1.7894405940529252 0.06930921474868512
86 1.7894405940529252 0.06563405852317139
87 1.7894405940529252 0.1129987120571947
88 1.7829112807307352 0.07716354326968933
89 1.7829112807307352 0.2822739242762733
90 1.7947751770032367 0.2897277817406305
91 2.053480616338808 0.2953684139441606
92 2.053480616338808 0.2963595635153453
93 1.829954359268447 0.4521614906817355
94 1.7947751770032367 0.1122237639881849
95 1.7947751770032367 0.2118842889059689
96 1.7947751770032367 0.053151949838918355
97 1.7947751770032367 0.026845471426657497
98 1.6543828974629788 0.20457212165601923
99 1.6702477966033398 0.1681886720437511
100 2.053480616338808 0.3625992106552349
101 1.6543828974629788 0.37232372087489907
102 1.6797359597024537 0.40742121476015747
103 1.6797359597024537 0.04148427482165218
104 1.6797359597024537 0.041503413104981846
105 1.829954359268447 0.10623353532139745
106 1.829954359268447 0.12423021161826675
107 1.6541073553097958 0.29527722219572894
108 2.053480616338808 0.2396156241929699
109 1.6849659687199632 0.29691358652099886
110 1.7894405940529252 0.2119243055630804
111 1.7894405940529252 0.14996645406711728
112 1.7894405940529252 0.08289990848889595
}\datatable

%0.3915357659385303
%0.274637251011996
%0.3045327281221439
%0.29970576961695417
%0.2846488670799311
%0.2659836794443669
%0.24918385911642843

\pgfplotsset{/dummy/workaround/.style={/pgfplots/axis on top}}

\node[scale=1] (kth_cr) at (5.7,-0.19)
{
\begin{tikzpicture}
  \begin{axis}
[
        xmin=1,
        xmax=105,
        ymax=120,
        %ymode=log,
        width=11cm,
        height =3.5cm,
        axis y line=center,
        axis x line=bottom,
        scaled y ticks=false,
        yticklabel style={
        /pgf/number format/fixed,
        /pgf/number format/precision=5
      },
        xlabel style={below right},
        ylabel style={above left},
        axis line style={-{Latex[length=2mm]}},
        smooth,
        legend style={at={(0.95,0.8)}},
        legend columns=1,
        legend style={
          /tikz/every node/.style={anchor=west},
          draw=none,
            /tikz/column 2/.style={
                column sep=5pt,
              }
              }
              ]
            \addplot[Black,name path=l1, thick, dashed] table [x index=0, y index=1, domain=0:1] {\datatable};              
            \addplot[RoyalAzure,name path=l1, thick] table [x index=0, y index=2, domain=0:1] {\datatable};
%mark=x, mark repeat=0
%              \addplot[Red,name path=l1, thick] table [x index=0, y index=1, domain=0:1] {\solvefour};
\legend{Theoretical bound [cf. \propref{prop:aggregation_bound}], Actual approximation error}
\end{axis}
\node[inner sep=0pt,align=center, scale=1, rotate=0, opacity=1] (obs) at (2.32,1.9)
{
  $\norm{\overline{J}^{\star}-\Tilde{J}}$ (approximation error)
};
\node[inner sep=0pt,align=center, scale=1, rotate=0, opacity=1] (obs) at (4.56,-0.8)
{
  Quantization resolution $r$
};
\end{tikzpicture}
};
\end{tikzpicture}
  }
  \caption{Comparison between the theoretical error bound in \propref{prop:aggregation_bound} and the actual error of the cost function approximation $\Tilde{J}$ [cf.~(\ref{eq:approximation_1})] for the illustrative example with $N=1$ and varying quantization resolutions $r$; cf.~(\ref{eq:aggregate_belief_space}).}
  \label{fig:rep21}
\end{figure}

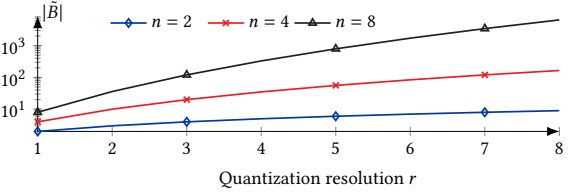
\begin{figure}
  \centering
%\tikzsetnextfilename{rep20}   
  \scalebox{0.8}{
    \begin{tikzpicture}

\pgfplotstableread{
0 1
1 2
2 3
3 4
4 5
5 6
6 7
7 8
8 9
}\statetwo

\pgfplotstableread{
0 1
1 4
2 10
3 20
4 35
5 56
6 84
7 120
8 165
}\statefour

\pgfplotstableread{
0 1
1 8
2 36
3 120
4 330
5 792
6 1716
7 3432
8 6435
}\stateeight

\pgfplotstableread{
1 6.12
2 6.12
3 4.12
4 4.62
5 4.01
6 3.93
7 3.45
8 3.33
9 3.04
10 2.85
11 2.73
12 2.5
13 2.44
14 2.28
15 2.25
16 2.14
17 2.03
18 1.96
19 1.87
20 1.83
21 1.53
22 1.25
23 1.26
24 0.96
25 0.9
26 0.73
27 0.73
28 0.62
29 0.62
30 0.57
31 0.53
32 0.54
33 0.46
34 0.48
35 0.4
36 0.41
37 0.42
38 0.4
39 0.41
40 0.35
41 0.36
42 0.35
43 0.35
44 0.35
45 0.35
46 0.37
47 0.36
48 0.35
49 0.33
50 0.33
51 0.3
52 0.29
53 0.3
54 0.3
55 0.31
56 0.29
57 0.28
58 0.27
59 0.27
60 0.27
61 0.28
62 0.27
63 0.28
64 0.28
65 0.28
66 0.28
67 0.26
68 0.27
69 0.27
70 0.28
71 0.29
72 0.29
73 0.29
74 0.28
75 0.28
76 0.28
77 0.28
78 0.28
79 0.28
80 0.27
81 0.27
82 0.28
83 0.27
84 0.28
85 0.27
86 0.26
87 0.26
88 0.26
89 0.27
90 0.26
91 0.26
92 0.26
93 0.26
94 0.26
95 0.25
96 0.25
97 0.25
98 0.25
99 0.25
100 0.26
101 0.26
102 0.26
103 0.26
104 0.25
105 0.25
106 0.25
107 0.26
108 0.26
109 0.26
110 0.26
111 0.26
112 0.26
113 0.25
114 0.24
115 0.24
116 0.24
117 0.25
118 0.25
119 0.25
120 0.24
}\solvetwo

\pgfplotstableread{
1 10.69
2 6.77
3 4.77
4 3.92
5 3.55
6 3.09
7 2.78
8 2.59
9 2.31
10 2.19
11 2.01
12 1.85
13 1.76
14 1.68
15 1.62
16 1.56
17 1.46
18 1.43
19 1.38
20 1.33
21 0.81
22 0.69
23 0.68
24 0.64
25 0.55
26 0.54
27 0.51
28 0.46
29 0.48
30 0.44
31 0.45
32 0.47
33 0.46
34 0.46
35 0.44
36 0.42
37 0.44
38 0.42
39 0.4
40 0.4
41 0.41
42 0.42
43 0.4
44 0.39
45 0.4
46 0.4
47 0.39
48 0.42
49 0.41
50 0.41
51 0.41
52 0.41
53 0.4
54 0.39
55 0.39
56 0.4
57 0.4
58 0.4
59 0.41
60 0.41
61 0.42
62 0.42
63 0.41
64 0.41
65 0.41
66 0.4
67 0.4
68 0.39
69 0.39
70 0.39
71 0.38
72 0.39
73 0.39
74 0.38
75 0.39
76 0.39
77 0.39
78 0.39
79 0.38
80 0.38
81 0.37
82 0.37
83 0.38
84 0.38
85 0.38
86 0.39
87 0.38
88 0.39
89 0.39
90 0.39
91 0.39
92 0.38
93 0.38
94 0.38
95 0.38
96 0.38
97 0.38
98 0.38
99 0.38
100 0.38
101 0.38
102 0.39
103 0.38
104 0.38
105 0.38
106 0.38
107 0.38
108 0.38
109 0.37
110 0.37
111 0.37
112 0.38
113 0.38
114 0.38
115 0.38
116 0.38
117 0.38
118 0.38
119 0.38
120 0.38
}\solvefour

\pgfplotstableread{
1 8.03
2 5.93
3 4.53
4 3.61
5 3.06
6 2.6
7 2.26
8 2.08
9 1.94
10 1.77
11 1.68
12 1.55
13 1.43
14 1.36
15 1.32
16 1.26
17 1.22
18 1.18
19 1.12
20 1.09
21 0.71
22 0.54
23 0.48
24 0.47
25 0.44
26 0.45
27 0.47
28 0.44
29 0.43
30 0.43
31 0.41
32 0.42
33 0.45
34 0.44
35 0.42
36 0.42
37 0.41
38 0.41
39 0.42
40 0.42
41 0.41
42 0.41
43 0.4
44 0.4
45 0.4
46 0.4
47 0.4
48 0.4
49 0.4
50 0.4
51 0.41
52 0.41
53 0.41
54 0.41
55 0.4
56 0.41
57 0.4
58 0.4
59 0.4
60 0.4
61 0.4
62 0.4
63 0.4
64 0.4
65 0.4
66 0.4
67 0.39
68 0.39
69 0.39
70 0.39
71 0.39
72 0.39
73 0.38
74 0.39
75 0.39
76 0.38
77 0.39
78 0.39
79 0.39
80 0.39
81 0.39
82 0.39
83 0.39
84 0.39
85 0.39
86 0.39
87 0.39
88 0.39
89 0.39
90 0.39
91 0.39
92 0.39
93 0.39
94 0.39
95 0.39
96 0.39
97 0.39
98 0.39
99 0.39
100 0.39
101 0.39
102 0.38
103 0.38
104 0.38
105 0.38
106 0.38
107 0.38
108 0.38
109 0.38
110 0.38
111 0.38
112 0.38
113 0.38
114 0.38
115 0.38
116 0.38
117 0.37
118 0.37
119 0.37
120 0.37
}\solveeight

\pgfplotstableread{
1 0.002092123031616211
2 0.003713846206665039
3 0.0038819313049316406
4 0.00777125358581543
5 0.011507034301757812
6 0.016611099243164062
7 0.02157282829284668
8 0.029363155364990234
9 0.03541398048400879
10 0.041932106018066406
}\vione

\pgfplotstableread{
1 0.1882920265197754
2 1.1791269779205322
3 4.848140001296997
4 16.423473834991455
5 47.63405084609985
6 126.22407865524292
7 289.82488918304443
8 645.9581429958344
9 1358.296599149704
}\vitwo

\pgfplotstableread{
1 0.8174553979430499
2 0.5277080386539204
3 0.4854515188566065
4 0.3271052118603257
5 0.21198936265286528
6 0.25357901350799983
7 0.22667558802944976
8 0.08996632869685334
9 0.20500989335070316
10 0.23030856291524912
11 0.21220307801542299
12 0.09035471292412017
13 0.16106149077505214
14 0.09563208319555039
15 0.07674626787741051
16 0.012792905270250843
17 0.07727795155661105
18 0.05024301265141917
19 0.026530299756261616
20 0.029875320741394262
21 0.02975385508367634
22 0.01108851154605639
23 0.000229497966178549
24 0.00935816225852154
25 0.008618224201830671
26 0.019104638089423698
27 0.046337821616816266
28 0.04541117345723611
29 0.015075318910745678
30 0.003344378758808886
31 0.010422806476308583
32 0.012655280641803185
33 0.01377959272979934
34 0.03074515584070675
35 0.010896367663508294
36 0.011058869433781537
37 0.016834969799636917
38 0.0521375981462969
39 0.05982065321407434
40 0.047564151673513494
41 0.07590018473852442
42 0.09230291773725319
43 0.05995467466802837
44 0.08983198287772234
45 0.08342883936188303
46 0.07944728812603918
47 0.07039239540798903
48 0.09259640119572113
49 0.05975773867719164
50 0.06595955715967924
}\policyone

\pgfplotstableread{
1 1.2583676643087585
2 0.7532216025677129
3 0.638458127809979
4 0.533530357052362
5 0.1507599166845921
6 0.06256073653859406
7 0.14183776087532485
8 0.131100091
9 0.1678192015342
10 0.16125102110
}\policytwo

\pgfplotstableread{
1 3.2019309997558594
2 3.339353084564209
3 3.1998019218444824
4 3.3643980026245117
5 3.187494993209839
6 3.3815529346466064
7 3.2713332176208496
8 3.269848108291626
9 3.3910391330718994
10 3.4765219688415527
}\policyevalone

\pgfplotstableread{
1 3.709481954574585
2 3.9555530548095703
3 3.9309158325195312
4 4.086724758148193
5 4.755553960800171
6 4.532117128372192
7 4.580238103866577
8 5.3591601848602295
9 5.988803148269653
10 6.23109192851254
}\policyevaltwo

\pgfplotstableread{
1 0.729
2 0.486
3 0.585
4 0.585
5 0.486
6 0.486
7 0.484
8 0.5
9 0.486
10 0.486
11 0.483
12 0.484
13 0.5
14 0.486
15 0.483
16 0.484
17 0.484
18 0.486
19 0.483
20 0.484
21 0.484
22 0.5
23 0.484
24 0.483
25 0.484
26 0.484
27 0.484
28 0.483
29 0.484
30 0.484
31 0.484
32 0.483
33 0.483
34 0.484
35 0.484
36 0.483
37 0.483
38 0.484
39 0.484
40 0.483
41 0.483
42 0.483
43 0.484
44 0.484
45 0.483
46 0.483
47 0.484
48 0.484
49 0.483
50 0.483
}\basecost

\pgfplotstableread{
1 0.484
2 0.484
3 0.483
4 0.483
5 0.484
6 0.484
7 0.482
8 0.483
9 0.484
10 0.484
11 0.482
12 0.482
13 0.483
14 0.484
15 0.482
16 0.482
17 0.482
18 0.484
19 0.482
20 0.482
21 0.482
22 0.483
23 0.483
24 0.482
25 0.482
26 0.482
27 0.483
28 0.482
29 0.482
30 0.482
31 0.482
32 0.482
33 0.482
34 0.482
35 0.482
36 0.482
37 0.482
38 0.482
39 0.482
40 0.482
41 0.482
42 0.482
43 0.482
44 0.483
45 0.482
46 0.482
47 0.482
48 0.483
49 0.482
50 0.482
}\rolloutcost

\pgfplotstableread{
1 0.484
2 0.482
3 0.483
4 0.483
5 0.482
6 0.482
7 0.482
8 0.482
9 0.482
10 0.482
11 0.482
12 0.482
13 0.482
14 0.482
15 0.482
16 0.482
17 0.482
18 0.482
19 0.482
20 0.482
21 0.482
22 0.482
23 0.482
24 0.482
25 0.482
26 0.482
27 0.482
28 0.482
29 0.482
30 0.482
31 0.482
32 0.482
33 0.482
34 0.482
35 0.482
36 0.482
37 0.482
38 0.482
39 0.482
40 0.482
41 0.482
42 0.482
43 0.482
44 0.482
45 0.482
46 0.482
47 0.482
48 0.482
49 0.482
50 0.482
}\rollouttwocost

\pgfplotstableread{
1 0.484
2 0.484
3 0.483
4 0.483
5 0.484
6 0.484
7 0.482
8 0.483
9 0.484
10 0.484
11 0.482
12 0.482
13 0.483
14 0.484
15 0.482
}\truncatedrolloutcost

\pgfplotstableread{
1 0.485
2 0.482
3 0.483
4 0.483
5 0.482
6 0.482
7 0.482
8 0.482
9 0.482
10 0.482
11 0.482
12 0.482
13 0.482
14 0.482
15 0.482
16 0.482
17 0.482
18 0.482
19 0.482
20 0.482
}\truncatedrolloutcosttwo

\pgfplotstableread{
1 0.584
2 0.482
3 0.483
4 0.483
5 0.482
6 0.482
7 0.482
8 0.483
9 0.482
10 0.482
}\cerolloutcost

\pgfplotstableread{
1 0.485
2 0.482
3 0.483
4 0.483
5 0.482
6 0.482
7 0.482
8 0.483
9 0.482
10 0.482
}\cerolloutcosttwo

\pgfplotstableread{
1 0.486
2 0.483
3 0.483
4 0.483
5 0.484
6 0.483
7 0.484
8 0.483
9 0.483
10 0.483
11 0.483
12 0.483
13 0.483
14 0.483
15 0.483
}\mcrolloutcosttwo

\pgfplotstableread{
1 0.486
2 0.484
3 0.483
4 0.483
5 0.484
6 0.484
7 0.484
8 0.483
9 0.484
10 0.484
11 0.485
12 0.483
13 0.484
14 0.484
15 0.485
}\mcrolloutcost

%0.3915357659385303
%0.274637251011996
%0.3045327281221439
%0.29970576961695417
%0.2846488670799311
%0.2659836794443669
%0.24918385911642843

\pgfplotsset{/dummy/workaround/.style={/pgfplots/axis on top}}

\node[scale=1] (kth_cr) at (0,0.065)
{
\begin{tikzpicture}
  \begin{axis}
[
        xmin=1,
        xmax=8,
        ymax=8500,
        ymode=log,
        width=10.25cm,
        height =3.5cm,
        axis y line=center,
        axis x line=bottom,
        scaled y ticks=false,
        yticklabel style={
        /pgf/number format/fixed,
        /pgf/number format/precision=5
      },
        xlabel style={below right},
        ylabel style={above left},
        axis line style={-{Latex[length=2mm]}},
%        smooth,
        legend style={at={(0.67,1.08)}},
        legend columns=3,
        legend style={
          draw=none,
            /tikz/column 2/.style={
                column sep=5pt,
              }
              }
              ]
              \addplot[RoyalAzure,name path=l1, thick, mark=diamond, mark repeat=2, samples=100] table [x index=0, y index=1, domain=0:1] {\statetwo};
              \addplot[Red,name path=l1, thick, mark=x, mark repeat=2, samples=100] table [x index=0, y index=1, domain=0:1] {\statefour};
              \addplot[Black,name path=l1, thick, mark=triangle, mark repeat=2, samples=100] table [x index=0, y index=1, domain=0:1] {\stateeight};
              \legend{$n=2$, $n=4$, $n=8$}
            \end{axis}
\node[inner sep=0pt,align=center, scale=1, rotate=0, opacity=1] (obs) at (0.29,2)
{
  $|\Tilde{B}|$
};
\node[inner sep=0pt,align=center, scale=1, rotate=0, opacity=1] (obs) at (4.7,-0.8)
{
  Quantization resolution $r$
};
\end{tikzpicture}
};
\end{tikzpicture}
  }
  \caption{Number of representative beliefs [cf.~(\ref{eq:aggregate_belief_space})] in function of the quantization resolution; curves relate to state spaces of different sizes.}\label{fig:rep22}
\end{figure}

Next, we analyze how close the bound in \propref{prop:aggregation_bound} is to the actual approximation error, i.e., the difference $\norm{\tilde{J}-\overline{J}^{\star}}_{\infty}$. \Figref{fig:rep21} shows that the bound is not tight but becomes increasingly accurate as the resolution $r$ increases, as asserted in \propref{prop:aggregation_bound}. However, increasing $r$ also causes the number of representative beliefs to grow, which is illustrated in \figref{fig:rep22}. Hence, $r$ governs a trade-off between computational expedience and approximation error.

%\tikzexternaldisable
% \tikzexternalenable

%\tikzexternaldisable
% \tikzexternalenable

\begin{figure}
  \centering
%\tikzsetnextfilename{value_fun}     
  \scalebox{0.77}{
    \input{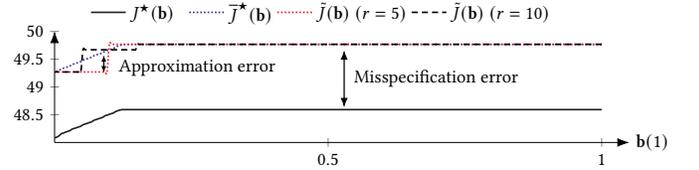}
  }
  \caption{Comparison between the optimal cost function $J^{\star}$, the optimal cost function in the conjectured model $\overline{J}^{\star}$, and the approximation $\Tilde{J}$ with varying quantization resolution $r$ [cf.~(\ref{eq:approximation_1})] for the illustrative example. The curves for the approximations are computed using $\overline{\bm{\theta}} = 0.5$ and $\bm{\theta}=0.2$. The number of system components in the example is $N=1$. Hence, the value $\mathbf{b}(1)$ on the x-axis indicates the belief of system compromise.}
  \label{fig:value_fun}
\end{figure}

\Figref{fig:value_fun} shows the structure of the optimal cost function $J^{\star}$, the conjectured optimal cost function $\overline{J}^{\star}$, and the cost function approximation $\Tilde{J}$; cf.~(\ref{eq:approximation_1}). We observe that $J^{\star}$ and $\overline{J}^{\star}$ have a similar structure but differ significantly in their values. Moreover, we observe that $\Tilde{J}$ is piece-wise constant, as expected from (\ref{eq:aggregation_mapping}).

%\tikzexternaldisable
% \tikzexternalenable

\section{Evaluation on the \textsc{cage-2} Benchmark}\label{sec:evaluation}
To compare our method with the state-of-the-art methods for computing incident response strategies, we apply it to the \textsc{cage-2} benchmark \citep{cage_challenge_2_announcement}. \textsc{cage-2} involves a networked system segmented into \textit{zones} with servers and workstations that run network services. The network topology of the \textsc{cage-2} system is shown in \figref{fig:cage_network}. The system provides services to clients through a gateway, which is also open to an attacker who aims to intrude on the system. These services generate a stream of network statistics, which are input to an incident response strategy $\pi$, which can take four \textit{actions} on each node: analyze it for a possible intrusion; start a decoy service; remove malware; and restore it to a secure state, which temporarily disrupts its service. Each service disruption and node compromise incurs a predefined cost; the problem is finding a response strategy that minimizes this cost. When formulated as a \textsc{pomdp}, \textsc{cage-2} has $145$ actions, over $10^{47}$ states, and over $10^{25}$ observations \citep{hammar2024optimaldefenderstrategiescage2}.

\subsubsection*{\textbf{Experimental setup}} We consider the standard \textsc{cage-2} setup with the \textsc{b-line} attacker and run the system for $100$ time steps. Due to the large state and observation spaces in \textsc{cage-2}, we employ the following approximations to instantiate \textsc{mobal}. First, we use the particle filter (\ref{eq:estimate_belief}) with $M=50$ particles to implement the belief estimator (\ref{eq:belief_estimator}). Second, we approximate the Bayesian update (\ref{eq:bayesian_learning}) using Monte-Carlo sampling. Third, since the dimension of the belief space $\mathcal{B}$ is larger than $10^{47}$, we reduce its dimension by only considering beliefs over the following state variables: the attacker's state (i.e., the attacker's location in the network), the attacker's next target, and the configuration of the decoys. These state features were originally proposed in \citep{tifs_25_HLALB} and lead to a belief space of dimension $427,500$, which we quantize with resolution $r=1$; cf.~(\ref{eq:aggregate_belief_space}).

\subsubsection*{\textbf{Evaluation scenarios}}
A central aspect of a response strategy in \textsc{cage-2} is the selection and placement of decoys, which are intended to mislead the attacker and divert attention away from vulnerable system components. The effectiveness of the decoys depends on the probability that an attacker will engage with them and the number of decoys available. \tableref{tab:decoys} lists the attack-probabilities in \textsc{cage-2} under different decoy configurations. As shown in the table, the probability of a successful attack decreases with the number of decoys. Current methods for \textsc{cage-2} assume that these probabilities are encoded in a simulator that can be used for numerical optimization of the response strategy. In practice, however, these probabilities cannot be known with certainty and can only be conjectured. For this reason, we consider both the standard \textsc{cage-2} scenario and a (more realistic) scenario in which the potential benefit of the decoys is unknown. The scenarios are detailed below.
\begin{enumerate}
\item \setword{\textsc{no misspecification}}{cage_scenario_1}: In this scenario, we consider the case where the model is correctly specified and known, i.e., $\rho_0(\bm{\theta}) = 1$, where $\bm{\theta}$ are the true parameters of the \textsc{cage-2} model. (The source code of \textsc{cage-2} is available in \citep{cage_challenge_2_announcement}.)
\item \setword{\textsc{misspecification}}{cage_scenario_2}: In this scenario, we consider the case where the model is misspecified, i.e., $\rho_0(\bm{\theta}) = 0$. We define the vector $\bm{\theta}$ to represent the conditional probability of a successful attack against a node given its decoy configuration. Accordingly, we define $\Theta$ to be a set of conjectures of these probabilities; see \tableref{tab:decoys}. We assume that all other parameters of the \textsc{cage-2} model are correctly specified.
\end{enumerate}

\subsubsection*{\textbf{Methods for comparison}} Over $35$ methods have been evaluated against the \textsc{cage-2} benchmark. We compare our method (\textsc{mobal}) against the state-of-the-art methods, namely: \textsc{cardiff} \citep{vyas2023automated} and \textsc{c-pomcp} \citep[Alg. 1]{hammar2024optimaldefenderstrategiescage2}. We also compare it against two baseline methods: \textsc{ppo} \citep[Alg. 1]{ppo} and \textsc{pomcp} \citep[Alg. 1]{pomcp}. For the \textsc{misspecification} scenario, we run these methods on a simulator of \textsc{cage-2} where all of the probability parameters listed in \tableref{tab:decoys} are fixed to $0.5$.

\subsubsection*{\textbf{Evaluation results}}
The evaluation results are summarized in \tableref{tab:evaluation_1}. In the \textsc{no misspecification scenario}, the results show that our method (\textsc{mobal}) performs slightly worse than the state-of-the-art (\textsc{c-pomcp} and \textsc{cardiff}), but performs significantly better than the baseline methods (\textsc{ppo} and \textsc{pomcp}). However, in the \textsc{misspecification} scenario, \textsc{mobal} significantly outperforms all other methods. We attribute the favorable performance of \textsc{mobal} to its ability to adapt the conjectured system model online based on system observations. By contrast, the existing methods assume a correctly specified model and cannot adapt it.
\begin{table}[H]
  \centering
  \scalebox{0.85}{
    \begin{tabular}{lll}
      \toprule
      \rowcolor{lightgray}
      \textit{Method} & \textit{Offline/Online compute time (min)} & \textit{Cost ($\downarrow$ better)} \\     
      \midrule
      \multicolumn{3}{c}{\textbf{No misspecification}} \\
\rowcolor{lightblue}      
      \textsc{mobal}  & $0/8.50$ & $15.19 \pm 0.82$ \\      
      \textsc{cardiff} \citep{vyas2023automated} & $300/0.01$ & $\bm{13.69} \pm 0.53$ \\
      \textsc{ppo} \citep{ppo} & $1000/0.01$ & $119.02 \pm 58.11$ \\
      \textsc{c-pomcp} \citep{hammar2024optimaldefenderstrategiescage2} & $0/0.50$ & $\bm{13.32} \pm 0.18$ \\
      \textsc{pomcp} \citep{pomcp} & $0/0.50$ & $29.51 \pm 2.00$ \\
      \midrule
      \multicolumn{3}{c}{\textbf{Misspecification}} \\
\rowcolor{lightblue}       
      \textsc{mobal}  & $0/8.50$ & $\bm{35.91} \pm 9.01$ \\      
      \textsc{cardiff} \citep{vyas2023automated} & $300/0.01$ & $94.28 \pm 33.27$ \\
      \textsc{ppo} \citep{ppo} & $1000/0.01$ & $124.38 \pm 55.49$ \\
      \textsc{c-pomcp} \citep{hammar2024optimaldefenderstrategiescage2} & $0/0.50$ & $92.71 \pm 27.67$ \\
      \textsc{pomcp} \citep{pomcp} & $0/0.50$ & $91.51 \pm 28.23$ \\
      \bottomrule
    \end{tabular}
  }
  \caption{Evaluation results on \textsc{cage-2}. Rows relate to different methods; columns indicate performance metrics. Results that are within the margin of statistical equivalence to the state-of-the-art are highlighted in bold. Numbers indicate the mean and the standard deviation from $100$ evaluations with $100$ time steps. The cost is calculated using \textsc{cage-2}'s internal cost function.}
  \label{tab:evaluation_1}
\end{table}
\subsubsection*{\textbf{Discussion of the evaluation results}}
The evaluation demonstrates the key benefit of our method (\textsc{mobal}), namely its robustness to model misspecification. While existing methods perform well when the system model is correctly specified, their reliance on a detailed model makes them brittle in practice where such models are unavailable. In contrast, \textsc{mobal} continuously adapts a conjecture about the model based on observed data, which allows it to respond effectively to attacks even under model misspecification.

\begin{table*}
  \centering
  \scalebox{0.85}{
    \begin{tabular}{l l l l l l l l l l}
      \toprule
      \rowcolor{lightgray}
      \textit{Node} & \textit{\textsc{smtp} decoy} & \textit{\textsc{tomcat} decoy} & \textit{\textsc{apache} decoy} & \textit{\textsc{ftp} decoy} & \textit{\textsc{femitter} decoy} & \textit{\textsc{smss} decoy} & \textit{\textsc{ssh} decoy} & \textit{Attack probability $\bm{\theta}$} &  \textit{Conjectures $\Theta$}  \\
      \midrule
      $1$ & \xmark & \xmark & \xmark & \xmark & \xmark & \xmark & \xmark & $1$ & $\{0, 0.5, 1\}$\\
      $1$ & \cmark & \xmark & \xmark & \xmark & \xmark & \xmark & \xmark & $0.25$ & $\{0, 0.5, 1\}$\\
      $1$ & \cmark & \cmark & \xmark & \xmark & \xmark & \xmark & \xmark & $0.12$ & $\{0, 0.5, 1\}$\\
      $1$ & \cmark & \cmark & \xmark & \cmark & \xmark & \xmark & \xmark & $0.08$ & $\{0, 0.5, 1\}$\\
      $1$ & \cmark & \cmark & \cmark & \cmark & \xmark & \xmark & \xmark & $0.08$ & $\{0, 0.5, 1\}$\\
      $2$ & \xmark & \xmark & \xmark & \xmark & \xmark & \xmark & \xmark & $1$ & $\{0, 0.5, 1\}$\\
      $2$ & \xmark & \xmark & \xmark & \xmark & \cmark & \xmark & \xmark & $0.25$ & $\{0, 0.5, 1\}$\\
      $3$ & \xmark & \xmark & \xmark & \xmark & \xmark & \xmark & \xmark & $1$ & $\{0, 0.5, 1\}$\\
      $3$ & \xmark & \xmark & \xmark & \xmark & \cmark & \xmark & \xmark & $0.25$ & $\{0, 0.5, 1\}$\\
      $7$ & \xmark & \xmark & \xmark & \xmark & \xmark & \xmark & \xmark & $1$ & $\{0, 0.5, 1\}$\\
      $7$ & \cmark & \xmark & \xmark & \xmark & \xmark & \xmark & \xmark & $0.25$ & $\{0, 0.5, 1\}$\\
      $7$ & \cmark & \xmark & \cmark & \xmark & \xmark & \xmark & \xmark & $0.13$ & $\{0, 0.5, 1\}$\\
      $7$ & \cmark & \cmark & \cmark & \xmark & \xmark & \xmark & \xmark & $0.08$ & $\{0, 0.5, 1\}$\\
      $7$ & \cmark & \cmark & \cmark & \cmark & \xmark & \xmark & \xmark & $0.08$ & $\{0, 0.5, 1\}$\\
      $9$ & \xmark & \xmark & \xmark & \xmark & \xmark & \xmark & \xmark & $1$ & $\{0, 0.5, 1\}$\\
      $9$ & \xmark & \xmark & \cmark & \xmark & \xmark & \xmark & \xmark & $0.09$ & $\{0, 0.5, 1\}$\\
      $9$ & \xmark & \cmark & \cmark & \xmark & \xmark & \xmark & \xmark & $0.08$ & $\{0, 0.5, 1\}$\\
      $9$ & \xmark & \cmark & \cmark & \xmark & \xmark & \cmark & \xmark & $0.08$ & $\{0, 0.5, 1\}$\\
      $9$ & \xmark & \cmark & \cmark & \xmark & \xmark & \cmark & \cmark & $0.08$ & $\{0, 0.5, 1\}$\\
      $10$ & \xmark & \xmark & \xmark & \xmark & \xmark & \xmark & \xmark & $1$ & $\{0, 0.5, 1\}$\\
      $10$ & \xmark & \xmark & \xmark & \xmark & \cmark & \xmark & \xmark & $0.25$ & $\{0, 0.5, 1\}$\\
      $10$ & \xmark & \cmark & \xmark & \xmark & \cmark & \xmark & \xmark & $0.17$ & $\{0, 0.5, 1\}$\\
      $10$ & \xmark & \cmark & \cmark & \xmark & \cmark & \xmark & \xmark & $0.12$ & $\{0, 0.5, 1\}$\\
      $10$ & \xmark & \cmark & \cmark & \xmark & \cmark & \xmark & \cmark & $0.10$ & $\{0, 0.5, 1\}$\\
      $11$ & \xmark & \xmark & \xmark & \xmark & \xmark & \xmark & \xmark & $1$ & $\{0, 0.5, 1\}$\\
      $11$ & \xmark & \xmark & \xmark & \cmark & \xmark & \xmark & \xmark & $1$ & $\{0, 0.5, 1\}$\\
      $11$ & \xmark & \xmark & \xmark & \cmark & \xmark & \xmark & \cmark & $0.09$ & $\{0, 0.5, 1\}$\\
      $12$ & \xmark & \xmark & \xmark & \xmark & \xmark & \xmark & \xmark & $1$ & $\{0, 0.5, 1\}$\\                                          
      \bottomrule
    \end{tabular}
  }
  \caption{Decoy configurations and attack probabilities for the \textsc{cage-2} system \citep{cage_challenge_2_announcement}. The node identifiers correspond to the identifiers shown in \figref{fig:cage_network}. The last column indicates the conjectured attack probabilities by our method (\textsc{mobal}). While the \textsc{cage-2} system includes $12$ nodes (including the clients and the defender), only a subset of them are amenable to host decoys, which is why not all nodes are listed in the table. Moreover, different types of decoys are compatible with different types of nodes, which is why not all decoy configurations are listed in the table; see \citep{cage_challenge_2_announcement} for details.}
  \label{tab:decoys}
\end{table*}

\section{Related Work}\label{sec:related_work}
Since the early 1980s, there has been a broad interest in automating security functions, especially in intrusion detection and incident response \citep{anderson1980monitoring}. Traditional methods for incident response rely on static rules that map infrastructure statistics to response actions \citep{wazuh,playbook_response}. The main drawback of these methods is their dependence on domain experts to configure the rules, a process that is both labor-intensive and costly. Substantial effort has been devoted to addressing this limitation by developing methods for \textit{automatically} computing effective incident response strategies. Three predominant approaches have emerged from this research: control-theoretic, reinforcement learning, and game-theoretic approaches.

\subsubsection*{\textbf{Control theory for automated incident response}}
Control theory provides a well-established mathematical framework for studying automatic systems. Therefore, it provides a foundational theory for automated incident response. Previous works that apply control theory to incident response in \textsc{it} systems include: \citep{7568529,miehling_attack_graph,dsn24_hammar_stadler}, all of which model incident response as the problem of controlling a discrete-time dynamical system and obtain optimal strategies through dynamic programming techniques.

\subsubsection*{\textbf{Reinforcement learning for automated incident response}}
Reinforcement learning has emerged as a promising approach to approximate optimal control strategies in scenarios where dynamic programming is not applicable, and fundamental breakthroughs demonstrated by systems like \textsc{alpha-go} \citep{deepmind_2} have inspired researchers to study reinforcement learning to automate security functions. Three early papers: \citep{rl_seminal}, \citep{rl_seminal_3}, and \citep{rl_seminal_2} analyze incident response and apply traditional reinforcement learning algorithms. They have catalyzed much follow-up research \citep{10.1145/3605764.3623986,10.1145/3605764.3623916,deep_rl_cyber_sec, kurt_rl, deep_air,tabular_Q_andy, li2024conjectural,rl_tnsm_24,10.1007/978-3-030-01554-1_9, hammar2025incidentresponseplanningusing}. These works show that \textit{deep} reinforcement learning is a scalable technique for approximating optimal response strategies. However, such methods often lack convergence guarantees and rely on efficient simulators for training.
\subsubsection*{\textbf{Game theory for automated incident response}}
Game theory stands out from control theory and reinforcement learning by focusing on decision-makers that \textit{reason strategically} about the opponents' behavior. The formulation of incident response as a game can be traced back to the early 2000s with works such as \citep{levente_old} and \citep{altman_jamming_1}. In addition to these early pioneers, numerous researchers have contributed to this line of research in the last two decades; see e.g., \citep{nework_security_alpcan, 7347426,9559403,apt_rl_simulation, r1_ref4}. These works study various aspects of security games, including the existence, uniqueness, and structure of equilibria, as well as computational methods. However, most of them are based on abstract models and how they generalize to complex systems like \textsc{cage-2} is unproven.

\subsubsection*{\textbf{Comparison with this paper}}
The main difference between this paper and the works referenced above is that we propose a method for \textit{online} learning of incident response strategies under \textit{model misspecification}. By contrast, virtually all referenced works are \textit{offline} methods that assume access to a \textit{correctly specified} system model or simulator. The advantage of our method (\textsc{mobal}) is that it applies to a much broader class of practical use cases.

The only existing method for response planning that manages model misspecification in a principled way is our earlier work \citep{10955193}, which has influenced aspects of \textsc{mobal}. However, the method proposed in \citep{10955193} is designed for a small-scale game-theoretic setting, whereas we focus on a large-scale \textsc{pomdp} setting. The benefit of our approach is that it allows us to compare \textsc{mobal} against the state-of-the-art on the \textsc{cage-2} benchmark. Another fundamental difference between \textsc{mobal} and the method proposed in \citep{10955193} is the computational approach. Whereas we compute response strategies based on model quantization and dynamic programming, the method in \citep{10955193} uses lookahead optimization and rollout \citep{bertsekas2024reinforcement}.

\section{Practical Considerations}
The practical deployment of \textsc{mobal} depends on the characteristics of the target environment, such as network topology, system size, and response time requirements. While our experimental evaluation in this paper is focused on \textsc{it} systems, our problem formulation is general and can be instantiated for a broad range of operational contexts, including on-premises, cloud-based, hybrid, and operational technology (\textsc{ot}) systems. Our \textsc{pomdp} model [cf. \sectionref{sec:formalization}] treats the environment as a set of partially observable states, actions, and observations, without imposing restrictions on the physical infrastructure. For example, in \textsc{it} systems, the observation $o$ may represent alerts from an intrusion detection system. Similarly, in \textsc{ot} systems, $o$ could capture sensor readings or control system alarms.

\section{Conclusion}\label{sec:conclusion}
Effective incident response often requires quick decisions based on partial (and possibly misleading) indicators of compromise. In this paper, we address this challenge by designing a method for incident response planning that explicitly accounts for model misspecification, which we call \textsc{mobal}: \underline{M}isspecified \underline{O}nline \underline{Ba}yesian \underline{L}earning. Our method starts from a potentially inaccurate model conjecture and continuously adapts it using Bayesian updates informed by system observations. To compute effective responses online, we quantize this conjecture at each time step into a finite Markov model, which enables efficient response planning via dynamic programming. We establish theoretical guarantees for convergence and derive bounds that quantify the effects of model misspecification and quantization. Experiments on the \textsc{cage-2} benchmark show that our method offers substantial improvements in robustness to model misspecification compared to the current state-of-the-art methods.

\subsubsection*{\textbf{Future work}} While we have evaluated \textsc{mobal} on the \textsc{cage-2} benchmark, testing it in additional environments is an important next step. Furthermore, the current online computational time of \textsc{mobal} is around 8.5 minutes per time step. Though this planning time is acceptable in many contexts, it may be prohibitive for time-critical incident response scenarios. Future research should therefore investigate ways to reduce the planning time. To this end, a promising approach is to combine \textsc{mobal} with offline computations.

\section*{Acknowledgments}
This research is supported by the Swedish Research Council under contract 2024-06436.

\bibliographystyle{ACM-Reference-Format}
\bibliography{references}

\appendix

\section{Experimental Setup}\label{appendix:hyperparameters}
All computations are performed on an \textsc{m2}-ultra processor. The hyperparameters are listed in \tableref{tab:hyperparams}. Notation is explained in \tableref{tab:notation}. We use the implementation of \textsc{cardiff} described in \citep{vyas2023automated} and the implementation of \textsc{c-pomcp} described in \citep{hammar2024optimaldefenderstrategiescage2}. For \textsc{ppo}, we use the \textsc{stable-baselines} implementation \citep{stable-baselines3}. We set the hyperparameters for these methods to be the same as those used in \citep{hammar2024optimaldefenderstrategiescage2}. We identify the dynamics of the quantized \textsc{mdp} through simulations. We solve the quantized \textsc{mdp} using value iteration.

\begin{table}[!ht]
  \centering
  \scalebox{0.75}{
    \begin{tabular}{ll} \toprule
\rowcolor{lightgray}
    {\textit{Parameter(s)}} & {\textit{Values}} \\ \midrule
    Convergence threshold of value iteration & $0.1$.\\
     Number of particles $M$ & $50$ [cf. (\ref{eq:estimate_belief})] \\
     Discount factor $\gamma$ & $0.99$ [cf. (\ref{eq:objective})]\\
    \bottomrule\\
  \end{tabular}}
  \caption{Hyperparameters.}\label{tab:hyperparams}
\end{table}

\begin{table}
  \centering
  \scalebox{0.6}{
    \begin{tabular}{ll} \toprule
\rowcolor{lightgray}
      {\textit{Notation(s)}} & {\textit{Description}} \\ \midrule
      $\mathcal{S},\mathcal{O},\mathcal{A},\mathcal{B}$ & State, observation, action and belief spaces; cf. \sectionref{sec:formalization}.  \\
      $n$ & Number of states, i.e., $\mathcal{S}=\{1,2,\hdots,n\}$; cf. \sectionref{sec:formalization}.  \\      
      $s_t, o_t, a_t, \mathbf{b}_t$ & State, observation, action and belief at time $t$; cf. \sectionref{sec:formalization}.  \\
      $p_{ss^{\prime}}(a)$ & Probability of the transition $s \rightarrow s^{\prime}$ under action $a$; cf. \sectionref{sec:formalization}.  \\
      $z(o \mid s)$ & Probability of the observation $o$ in state $s$; cf. \sectionref{sec:formalization}.  \\
      $c(s,a)$ & Cost in state $s$ when taking action $a$; cf. \sectionref{sec:formalization}.  \\
      $\Theta, \rho_t$ & Set of parameter vectors and conjecture distribution; cf. \sectionref{sec:formalization} and (\ref{eq:bayesian_learning}).  \\
      $\bm{\theta},\overline{\bm{\theta}}$ & True and conjectured parameter vector; cf. \sectionref{sec:formalization}.  \\                  
      $\pi, \pi^{\star}, J^{\star}$ & Strategy, optimal strategy, and optimal cost function; cf. (\ref{eq:objective}).  \\
      $\mathbb{B}, \gamma$ & The belief estimator and discount factor; cf. (\ref{eq:belief_estimator}) and (\ref{eq:objective}).  \\
      $\widehat{\mathbf{b}}_t, M$ & Belief estimated through a particle filter and number of particles; cf. (\ref{eq:estimate_belief}).  \\
      $N$ & Number of system components in the illustrative example; cf. \sectionref{sec:example}.  \\            
      $P, K$ & Probability measure and discrepancy function; cf. \sectionref{sec:bayesian_learning} and (\ref{eq:consistent_conjecture_sets}).  \\
      $K^{\star}, \Theta^{\star}$ & The mininmal value of $K$ and set of consistent conjectures; cf. \propref{label:prop_consistency} and (\ref{eq:consistent_conjecture_sets}).  \\
      $\hat{c}$ & Cost function in the belief-\textsc{mdp}; cf. (\ref{eq:belief_cost}).\\
      $p_{\bm{\theta}}(\mathbf{b}^{\prime} \mid \mathbf{b}, a)$ & Probability of the transition in $\mathbf{b} \rightarrow \mathbf{b}^{\prime}$ in the belief-\textsc{mdp}; cf. \sectionref{sec:formalization}.\\
      $\hat{p}_{\overline{\bm{\theta}}}(\mathbf{b}^{\prime} \mid \mathbf{b}, a)$ & Probability of the transition in $\mathbf{b} \rightarrow \mathbf{b}^{\prime}$ in the (conjectured) quantized belief-\textsc{mdp}; cf. \sectionref{sec:quantization}.\\            
      $\Tilde{\mathcal{B}}, r$ & Set of representative beliefs and quantization resolution; cf. (\ref{eq:aggregate_belief_space}).\\
      $\Phi,\tilde{\mathbf{b}}$ & Nearest-neighbor mapping and representative belief; cf. (\ref{eq:aggregation_mapping}) and (\ref{eq:aggregate_belief_space}).\\
      $\mu^{\star},V^{\star}$ & Optimal strategy and cost function in the quantized \textsc{mdp}; cf. (\ref{eq:approximation_1}).\\
      $\overline{\pi}^{\star}, \overline{J}^{\star}$ & Optimal strategy and cost function in the \textsc{pomdp} defined by $\overline{\bm{\theta}}$; cf. \sectionref{sec:quantization}.\\      
      $\Tilde{J},\tilde{\pi}$ & Cost function approximation and strategy approximation; cf. (\ref{eq:approximation_1}).\\                  
      $S_{\tilde{\mathbf{b}}}$ & Set of beliefs that are mapped to $\tilde{\mathbf{b}}$ in the quantization; cf. \propref{prop:aggregation_bound}.\\
    \bottomrule\\
  \end{tabular}}
  \caption{Notation.}\label{tab:notation}
\end{table}

\section{Proof of  \Propref{label:prop_consistency}}\label{appendix:regularity}
\Propref{label:prop_consistency} holds under the following two assumptions.

\begin{assumption}[Well-defined prior and Bayesian learning]\label{assumption:bayes}\mbox{}\newline
\begin{enumerate}[(i)]
\vspace{-1em}
\item The set $\Theta$ is a compact subset of a Euclidean space.
\item The prior $\rho_0$ has full support, i.e, $\rho_0(\overline{\bm{\theta}}) > 0$ for all $\overline{\bm{\theta}} \in \Theta$.
\end{enumerate}
\end{assumption}

\begin{assumption}[Regularity conditions]\label{assumption:regularity}
For any given observation $o \in \mathcal{O}$ and parameter vector $\bm{\theta} \in \Theta$,
  \begin{enumerate}
  \item The mapping $\mathbf{b}\mapsto \ln P(o \mid \bm{\theta}, \mathbf{b}, a)$ is Lipschitz w.r.t. the Wasserstein-$1$ distance, and the Lipschitz constant is independent of the observation $o$, the vector $\bm{\theta}$, and the action $a$.
  \item The mapping $\bm{\theta}\mapsto \ln P(o \mid \bm{\theta}, \mathbf{b}, a)$ is continuous and there exists an integrable function $g_{\mathbf{b},a}(o)$ for all beliefs $\mathbf{b}\in \mathcal{B}$ and actions $a \in \mathcal{A}$ such that $|\ln\frac{P(o \mid \bm{\theta}, \mathbf{b}, a)}{P(o \mid \overline{\bm{\theta}}, \mathbf{b}, a)}|\leq g_{\mathbf{b},a}(o)$ for all parameter vectors $\overline{\bm{\theta}}\in \Theta$.
  \end{enumerate}
\end{assumption}
Due to page restrictions, we present only the main proof steps here. See our earlier work \citep[Thm. 3]{10955193} for technical details. To begin with, we invoke two lemmas from \citep[Lemma 8, 9]{10955193} that ensure the regularity of the belief space and the integrand in \eqref{eq:consistent-conjecture}.
%The proofs of the following lemmas are built on the regularity conditions in Assumption~\ref{assumption:regularity}.

\begin{lemma}[Compact Measure Space]
\label{lem:compact}
    The belief space $\mathcal{B}\subset \mathbb{R}^n$ is compact with the Euclidean distance $d(\cdot, \cdot)$ and its corresponding Borel probability measure space $\Delta(\mathcal{B})$ is also compact with metric Wasserstein-$p$ distance $d_{\mathcal{W}}(\cdot, \cdot)$.
\end{lemma}

\begin{lemma}[Continuity]
\label{lem:contin}
    $\Delta K(\bar{\bm{\theta}}, \nu)\triangleq K(\bar{\bm{\theta}},\nu)-K_\Theta^\star(\nu)$ is a continuous mapping defined over the product space $\mathcal{B}\times \Delta(\mathcal{B})$ with respect to the product metric of $d(\cdot, \cdot)$ and $d_\mathcal{W}(\cdot, \cdot)$. 
\end{lemma}

Finally, the following lemma clarifies the probability measure under which the almost-sure convergence holds \citep[Lemma 6]{10955193}.
\begin{lemma}
    Any sequence of incident response strategies given by our method induces a well-defined probability measure over the sequence of historical states, partial observations, actions, and conjectures through the Ionescu-Tulcea extension \citep{IonescuTulcea1949}.
\end{lemma}

We now address the proof of \propref{label:prop_consistency}. Given a conjecture $\overline{\bm{\theta}}$ and the true model $\bm{\theta}$, recursively applying \eqref{eq:bayesian_learning} gives 
\begin{align*}
    \rho_{t+1}(\overline{\bm{\theta}})&=\frac{\rho_0(\overline{\bm{\theta}})\prod_{\tau=1}^t P(o_\tau \mid \overline{\bm{\theta}}, \mathbf{b}_{\tau-1}, a_{\tau-1})}{\int_\Theta \rho_0({\bm{\theta}}') \prod_{\tau=1}^t P(o_\tau \mid {\bm{\theta}'}, \mathbf{b}_{\tau-1}, a_{\tau-1}) \mathrm{d} {\bm{\theta}}'}\\
    &=\frac{\rho_0(\overline{\bm{\theta}})\exp\left(\ln \left(\prod_{\tau=1}^t \frac{P(o_\tau \mid \overline{\bm{\theta}}, \mathbf{b}_{\tau-1}, a_{\tau-1})}{P(o_\tau \mid \bm{\theta}, \mathbf{b}_{\tau-1}, a_{\tau-1})}\right) \right)}{\int_\Theta \rho_0({\bm{\theta}}') \exp\left(\ln \left(\prod_{\tau=1}^t \frac{P(o_\tau \mid \bm{\theta}', \mathbf{b}_{\tau-1}, a_{\tau-1})}{P(o_\tau \mid \bm{\theta}, \mathbf{b}_{\tau-1}, a_{\tau-1})}\right) \right) \mathrm{d} {\bm{\theta}}'}\\
    &=\frac{\rho_0(\overline{\bm{\theta}})\exp(-t Z_{t}(\overline{\bm{\theta}}))}{\int_\Theta \rho_0({\bm{\theta}}') \exp(-t Z_t({\bm{\theta}}')) \mathrm{d} {\bm{\theta}}'},
\end{align*}
where
\begin{align*}
Z_t(\overline{\bm{\theta}})&\triangleq t^{-1}\sum_{\tau=1}^t \ln \left( \frac{P(o_\tau\mid \bm{\theta}, \mathbf{b}_{\tau-1}, a_{\tau-1})}{P(o_\tau \mid \overline{\bm{\theta}}, \mathbf{b}_{\tau-1}, a_{\tau-1})} \right).  
\end{align*}
Plugging the expression above into the left-hand side of \eqref{eq:consistent-conjecture} yields
\begin{align}
\label{eq:lhs}
\frac{\int_\Theta \Delta K(\overline{\bm{\theta}}, \nu_t)\exp(-t Z_t(\overline{\bm{\theta}}))\rho_{0}(\overline{\bm{\theta}})\mathrm{d}\overline{\bm{\theta}} }{\int_\Theta \rho_0({\bm{\theta}}') \exp(-t Z_t({\bm{\theta}}')) \mathrm{d} {\bm{\theta}}'}.
\end{align}
Given the structure of the numerator above, we can partition the set $\Theta$ into $\Theta^{+}_{ \epsilon}\triangleq \{\overline{\bm{\theta}}: \Delta K(\overline{\bm{\theta}}, \nu_t)\geq \epsilon\}$ and $\Theta^{-}_{ \epsilon/2}\triangleq \{\overline{\bm{\theta}}: \Delta K(\overline{\bm{\theta}}, \nu_t)\leq \epsilon/2\}$, and the complement set of $\Theta^{+}_{ \epsilon}\cup \Theta^{-}_{ \epsilon/2}$ for any $\epsilon>0$ and $\nu_t$. Using this partitioning, (\ref{eq:lhs}) admits the following upper bound
\begin{align}
\label{eq:upper-bound}
    &\frac{\int_\Theta \Delta K(\overline{\bm{\theta}}, \nu_t)\exp(-t Z_t(\overline{\bm{\theta}}))\rho_0(\overline{\bm{\theta}})\mathrm{d}\overline{\bm{\theta}} }{\int_\Theta \rho_0({\bm{\theta}}') \exp(-t Z_t({\bm{\theta}}')) \mathrm{d} {\bm{\theta}}'}\nonumber \\
    &\qquad  \leq \frac{\left(\int_{\Theta\setminus \Theta^{+}_\epsilon}+\int_{\Theta^{+}_\epsilon}\right) \Delta K(\overline{\bm{\theta}}, \nu_t)\exp(-t Z_t(\overline{\bm{\theta}}))\rho_0(\overline{\bm{\theta}})\mathrm{d}\overline{\bm{\theta}} }{\int_\Theta \rho_0({\bm{\theta}}') \exp(-t Z_t({\bm{\theta}}')) \mathrm{d} {\bm{\theta}}'}\nonumber \\
    & \qquad \leq \epsilon + \underbrace{\frac{\int_{\Theta^{+}_\epsilon} \Delta K(\overline{\bm{\theta}}, \nu_t)\exp(-t Z_t(\overline{\bm{\theta}}))\rho_0(\overline{\bm{\theta}})\mathrm{d}\overline{\bm{\theta}}}{\int_{\Theta_{\epsilon/2}^{-}} \rho_0({\bm{\theta}}') \exp(-t Z_t({\bm{\theta}}')) \mathrm{d} {\bm{\theta}}'} }_{(*)}.
\end{align}
It suffices to prove that $(*)$ converges to zero for any $\epsilon>0$. 

Multiply both the numerator and denominator by $\exp(tK_\Theta^{\star}(\nu_t))$ in $(*)$, and we obtain 
\begin{equation*}
    (*)=\frac{\int_{\Theta^{+}_\epsilon} \Delta K(\overline{\bm{\theta}}, \nu_t)\exp(-t( Z_t(\overline{\bm{\theta}})-K^\star_{\Theta}(\nu_t)))\rho_0(\overline{\bm{\theta}})\mathrm{d}\overline{\bm{\theta}}}{\int_{\Theta_{\epsilon/2}^{-}} \rho_0({\bm{\theta}}') \exp(-t ( Z_t({\bm{\theta}}')-K^\star_{\Theta}(\nu_t))) \mathrm{d} {\bm{\theta}}'}.
\end{equation*}
According to \citep[Lemma 7]{10955193}, 
$\lim_{t\rightarrow\infty} |Z_t(\overline{\bm{\theta}})- K(\overline{\bm{\theta}},\nu_t)
    |=0$, almost surely,
which implies that asymptotically, $Z_t(\overline{\bm{\theta}})-K_\Theta^\star(\nu_t)$ is equivalent to $K(\overline{\bm{\theta}}, \nu_t)-K^\star_{\Theta}(\nu_t)$, and hence, 
\begin{align*}
    (*)&=\frac{\int_{\Theta^{+}_\epsilon} \Delta K(\overline{\bm{\theta}}, \nu_t)\exp(-t\Delta K(\overline{\bm{\theta}},\nu_t))\rho_0(\overline{\bm{\theta}})\mathrm{d}\overline{\bm{\theta}}}{\int_{\Theta_{\epsilon/2}^{-}}  \exp(-t\Delta K({\bm{\theta}}',\nu_t)) \rho_0({\bm{\theta}}')\mathrm{d} {\bm{\theta}}'}\\
    &\leq \frac{\epsilon e^{-t\epsilon}}{\int_{\Theta^{-}_{\epsilon/2}} e^{-\epsilon t/2} \rho_0({\bm{\theta}}')\mathrm{d} {\bm{\theta}}'}.
\end{align*}
Therefore, to prove that the upper bound of the left-hand side of \eqref{eq:consistent-conjecture} vanishes, we need to show that $\rho_0(\Theta_{\epsilon/2}^{-})\triangleq \int_{\Theta^{-}_{\epsilon/2}} \rho_0({\bm{\theta}}')\mathrm{d} {\bm{\theta}'}$ is strictly greater than zero for every $t$, for which the compactness result in Lemma~\ref{lem:compact} becomes helpful. 

Compactness of the parameter set $\Theta$ and the continuity of $\Delta K(\overline{\bm{\theta}}, \nu)$ imply its uniform continuity.  Berge's maximum theorem implies that $\Theta^\star(\nu)$ is non-empty \citep[Thm. 17.31]{aliprantis2006}. Therefore, for any $\overline{\bm{\theta}}_v\in \Theta^\star(\nu)$, there exist  $\overline{\bm{\theta}}\in \Theta$, $\nu'\in \Delta(\mathcal{B})$, and $\delta_m$ such that when $d(\overline{\bm{\theta}}_\nu, \overline{\bm{\theta}}')< \delta_m$ and $d_{\mathcal{W}}(\nu, \nu')<\delta_m$, $\Delta K(\overline{\bm{\theta}}',\nu')=\Delta K(\overline{\bm{\theta}}',\nu')- \Delta K(\overline{\bm{\theta}}_\nu, \nu)\leq m$, where the equality follows because $\overline{\bm{\theta}}_\nu\in \Theta^\star(\nu)$ implies $\Delta K(\overline{\bm{\theta}}_\nu, \nu)=0$. As a result, for any $\nu\in \Delta(\mathcal{B})$ and $\nu'\in B(\nu, \delta_m)\triangleq \{\nu' \mid  d_\mathcal{W}(\nu,\nu')< \delta_m\}$,
\begin{align*}
\underbrace{\{\overline{\bm{\theta}}'\mid d(\overline{\bm{\theta}}',\overline{\bm{\theta}}_\nu)< \delta_m\}}_{\Theta_{\nu}(\delta_m)} \subset \underbrace{\{\overline{\bm{\theta}}'\mid \Delta K(\overline{\bm{\theta}}',\nu')\leq m \}}_{\Theta_{\nu^{\prime}}(m)}.
\end{align*}
Thus, for any $\nu$ and $\nu'\in B(\nu, \delta_m)$, $\rho_0(\Theta_{\nu^{\prime}}(m))\geq \rho_0(\delta_m))>0$, where the strict inequality follows because $\rho_0$ has full support (Assumption \ref{assumption:bayes}).

The set $\{B(\nu,\delta_m)\}_{\nu\in \Delta(\mathcal{B})}$ forms an open cover for a compact space, implying that there exists a finite subcover $\{B(\nu_i,\delta_m)\}_{i=1}^M$. As a consequence, $\nu'\in \Delta(\mathcal{B})$ belongs to some Wasserstein ball $B(\nu_i,\delta_m)$. Let $r \triangleq \min_{i}\rho_0(\Theta_{\nu_i}(\delta_m)) > 0$. We obtain
$
\rho_0(\Theta_{\nu^{\prime}}(m))\geq \rho_0(\Theta_{\nu_i}(\delta_m)) \geq r,
$
which yields $\rho_0(\Theta_{ {\epsilon}/{2}}^{-})\geq r>0$ with $m={\epsilon}/{2}$. \qed
\section{Proof of \Propref{prop:misspecification_bound}}\label{app:prop_2_proof}
The proof follows the same chain of reasoning as the proof of the simulation lemma in \citep{Kearns2002}. We start by expanding the difference $|\overline{J}^{\star}(\mathbf{b})-J^{\star}(\mathbf{b})|$ as follows.
\begin{align*}
&|\overline{J}^{\star}(\mathbf{b})-J^{\star}(\mathbf{b})| = \Biggl| \hat{c}(\mathbf{b}, a) + \gamma\sum_{\mathbf{b}^{\prime} \in \mathcal{B}} p_{\overline{\bm{\theta}}}(\mathbf{b}^{\prime} \mid \mathbf{b},a)\overline{J}^{\star}(\mathbf{b}^{\prime})  - \biggl(\\
&\quad\quad\quad\quad\quad\quad\quad\quad\quad\hat{c}(\mathbf{b},a) + \gamma\sum_{\mathbf{b}^{\prime} \in \mathcal{B}} p_{\bm{\theta}}(\mathbf{b}^{\prime} \mid \mathbf{b},a)J^{\star}(\mathbf{b}^{\prime})\biggr)\Biggr|  \\
&=  \Biggl| \gamma\sum_{\mathbf{b}^{\prime} \in \mathcal{B}} p_{\overline{\bm{\theta}}}(\mathbf{b}^{\prime} \mid \mathbf{b},a)\overline{J}^{\star}(\mathbf{b}^{\prime})  - \gamma\sum_{\mathbf{b}^{\prime} \in \mathcal{B}} p_{\bm{\theta}}(\mathbf{b}^{\prime} \mid \mathbf{b},a)J^{\star}(\mathbf{b}^{\prime})\Biggr|  \\
&=  \Biggl| \gamma\sum_{\mathbf{b}^{\prime} \in \mathcal{B}} p_{\overline{\bm{\theta}}}(\mathbf{b}^{\prime} \mid \mathbf{b},a)\overline{J}^{\star}(\mathbf{b}^{\prime})  - \gamma\sum_{\mathbf{b}^{\prime} \in \mathcal{B}} p_{\bm{\theta}}(\mathbf{b}^{\prime} \mid \mathbf{b},a)J^{\star}(\mathbf{b}^{\prime}) + \\
& \quad\quad\gamma\sum_{\mathbf{b}^{\prime} \in \mathcal{B}} p_{\overline{\bm{\theta}}}(\mathbf{b}^{\prime} \mid \mathbf{b},a)J^{\star}(\mathbf{b}^{\prime}) - \gamma\sum_{\mathbf{b}^{\prime} \in \mathcal{B}} p_{\overline{\bm{\theta}}}(\mathbf{b}^{\prime} \mid \mathbf{b},a)J^{\star}(\mathbf{b}^{\prime})\Biggr|\\
&=  \Biggl| \gamma\sum_{\mathbf{b}^{\prime} \in \mathcal{B}} p_{\overline{\bm{\theta}}}(\mathbf{b}^{\prime} \mid \mathbf{b},a)\left(\overline{J}^{\star}(\mathbf{b}^{\prime})-J^{\star}(\mathbf{b}^{\prime})\right)  + \\
&\quad\quad \gamma\sum_{\mathbf{b}^{\prime} \in \mathcal{B}} \left(p_{\overline{\bm{\theta}}}(\mathbf{b}^{\prime} \mid \mathbf{b},a)-p_{\bm{\theta}}(\mathbf{b}^{\prime} \mid \mathbf{b},a)\right)J^{\star}(\mathbf{b}^{\prime})\Biggr|\\
&\leq \gamma\norm{\overline{J}^{\star}-J^{\star}}_{\infty} + \Biggl|\gamma\sum_{\mathbf{b}^{\prime} \in \mathcal{B}} \left(p_{\overline{\bm{\theta}}}(\mathbf{b}^{\prime} \mid \mathbf{b},a)-p_{\bm{\theta}}(\mathbf{b}^{\prime} \mid \mathbf{b},a)\right)J^{\star}(\mathbf{b}^{\prime})\Biggr|\\
&\numleq{a} \gamma\norm{\overline{J}^{\star}-J^{\star}}_{\infty} + \gamma\sum_{\mathbf{b}^{\prime} \in \mathcal{B}} \biggl|\left(p_{\overline{\bm{\theta}}}(\mathbf{b}^{\prime} \mid \mathbf{b},a)-p_{\bm{\theta}}(\mathbf{b}^{\prime} \mid \mathbf{b},a)\right)\biggr|\frac{c_{\text{\textsc{max}}}}{1-\gamma}  \\
&\leq \gamma\norm{\overline{J}^{\star}-J^{\star}}_{\infty} + \frac{\gamma \alpha c_{\text{\textsc{max}}}}{1-\gamma},
\end{align*}
where $\hat{c}(\mathbf{b},a)$ is defined in (\ref{eq:belief_cost}) and (a) follows because $|J^{\star}(\mathbf{b})| \leq \sum_{t=0}^{\infty}\gamma^{t}c_{\text{\textsc{max}}} = \frac{c_{\text{\textsc{max}}}}{1-\gamma}$ and the fact that $|ab|=|a||b|$ (we use the triangle inequality to move the absolute value inside the sum). Since this upper bound holds for any belief state $\mathbf{b}$, we have
\begin{align*}
\norm{\overline{J}^{\star}-J^{\star}}_{\infty} &\leq \gamma\norm{\overline{J}^{\star}-J^{\star}}_{\infty} + \frac{\gamma \alpha c_{\text{\textsc{max}}}}{1-\gamma}\\
\implies \norm{\overline{J}^{\star}-J^{\star}}_{\infty}-\gamma\norm{\overline{J}^{\star}-J^{\star}}_{\infty}&\leq \frac{\gamma \alpha c_{\text{\textsc{max}}}}{1-\gamma}\\
\implies (1-\gamma)\norm{\overline{J}^{\star}-J^{\star}}_{\infty}&\leq \frac{\gamma \alpha c_{\text{\textsc{max}}}}{1-\gamma}  \\
\implies \norm{\overline{J}^{\star}-J^{\star}}_{\infty}&\leq \frac{\gamma \alpha c_{\text{\textsc{max}}}}{(1-\gamma)^2}. \qed
\end{align*}
\section{Proof of \Propref{prop:aggregation_bound}}
The result expressed in \propref{prop:aggregation_bound} was originally proven by Tsitsiklis and van Roy in \citep[Thm. 1]{Tsitsiklis1996}, and later generalized by Li et al. \citep[Prop. 3]{lihambert}. Variants of this proof are also presented by Bertsekas in \citep{bertsekas2018featurebasedaggregationdeepreinforcement,bertsekas2019reinforcement,bertsekas2024reinforcement}. As this result is well established, we omit the proof. 

\section{Proof of \Propref{prop:consistent}}
Our proof is based on the arguments outlined by Hammar et al. in \citep[Prop. 2]{tifs_25_HLALB}. See also the proofs by Saldi et al. in \citep[Thm. 2.2]{saldi2016} and Yu and Bertsekas \citep[Thm. 1]{10.5555/1036843.1036918} for extensions to non-finite \textsc{pomdp}s and \textsc{pomdp}s with the average-cost criterion.

It can be shown that the (conjectured) optimal cost function $\overline{J}^{\star}:\mathcal{B} \mapsto \mathbb{R}$ is uniformly continuous; see e.g., \citep[Prop. 2.1]{10.5555/1195372}. Fix an arbitrary scalar $\omega>0$. By uniform continuity, there exists a scalar $\delta > 0$ such that
\begin{align}\label{eq:uniform}
\norm{\mathbf{b}-\mathbf{b}^{\prime}}_{\infty} < \delta \implies |\overline{J}^{\star}(\mathbf{b})-\overline{J}^{\star}(\mathbf{b}^{\prime})| < \omega,
\end{align}
for all $\mathbf{b},\mathbf{b}^{\prime}\in \mathcal{B}$. The quantization in (\ref{eq:aggregate_belief_space}) partitions $\mathcal{B}$ into grid cells $S_{\tilde{\mathbf{b}}}$ with resolution $r \geq 1$; cf.~(\ref{eq:partitioning}). Further, (\ref{eq:aggregation_mapping}) implies that if $\mathbf{b} \in S_{\tilde{\mathbf{b}}}$, then 
\begin{align*}
\norm{\mathbf{b}-\tilde{\mathbf{b}}}_{\infty}=\min_{\tilde{\mathbf{b}}^{\prime} \in \Tilde{\mathcal{B}}}\norm{\mathbf{b}-\tilde{\mathbf{b}}^{\prime}}_{\infty}.
\end{align*}
Because each belief coordinate $\mathbf{b}(s)$ lies in $[0,1]$ and each representative belief coordinate $\tilde{\mathbf{b}}(s)$ equals $\frac{\beta_s}{r}$ for some $\beta_s\in \{0,\dots,r\}$ [cf.~(\ref{eq:aggregate_belief_space})], we have
\begin{align*}
\max_{\mathbf{b},\mathbf{b}^{\prime}\in S_{\tilde{\mathbf{b}}}}\norm{\mathbf{b}-\mathbf{b}^{\prime}}_{\infty} \leq \frac{2n}{r}, \qquad\text{for every } \tilde{\mathbf{b}} \in \Tilde{\mathcal{B}}.
\end{align*}
Choose any $r$ such that $\frac{1}{r} < \delta$. By (\ref{eq:uniform}), we have
\begin{align*}
|\overline{J}^{\star}(\mathbf{b})-\overline{J}^{\star}(\mathbf{b}^{\prime})| < \omega,  \quad\text{for all } \mathbf{b},\mathbf{b}^{\prime}\in S_{\tilde{\mathbf{b}}},\; \tilde{\mathbf{b}} \in \Tilde{\mathcal{B}}.
\end{align*}  
Because $\omega>0$ is arbitrary and there exists a large enough $r$ such that $\frac{1}{r} < \delta$ for any $\delta > 0$, we have
\begin{align*}
\lim_{r\to\infty}\max_{\tilde{\mathbf{b}} \in \Tilde{\mathcal{B}}}\max_{\mathbf{b},\mathbf{b}^{\prime}\in S_{\tilde{\mathbf{b}}}} |\overline{J}^{\star}(\mathbf{b})-\overline{J}^{\star}(\mathbf{b}^{\prime})| =0.
\end{align*}
Hence the constant $\epsilon$ in \propref{prop:aggregation_bound} diminishes as $r \rightarrow \infty$. Invoking the error bound in \propref{prop:aggregation_bound} completes the proof. \qed

\end{document}